\documentclass[journal]{IEEEtran}

\ifCLASSINFOpdf

\else

\fi

\usepackage{graphicx,wrapfig,fullpage,amsmath,hhline,epsfig,verbatim,url,amssymb,multicol,multirow,cite}
\usepackage{times,color,soul} 
\usepackage{amsthm,amssymb}
\usepackage[left=0.75in,top=0.70in,right=0.75in,bottom=0.80in]{geometry}
\usepackage{amsbsy,latexsym,mathrsfs,mathtools}

\usepackage{mathptmx} 
\DeclareMathAlphabet{\mathcal}{OMS}{cmsy}{m}{n}
\usepackage{bm}
\usepackage{float}
\usepackage{color,soul}
\usepackage{dsfont}
\usepackage{comment}
\usepackage{algorithm}
\usepackage{algorithmic}
\usepackage{psfrag}   
\usepackage{tabularx}

\usepackage{url}
\usepackage{hyperref}
\usepackage{setspace}
\usepackage[table]{xcolor}
\usepackage{paralist}
\usepackage{booktabs}
\usepackage{gensymb}
\usepackage{lineno}
\usepackage{flushend}
\newcommand{\YanCom}[1]{\textcolor{red}{#1}}

\begin{document}

\title{DRS-LIP: Linear Inverted Pendulum Model for Legged Locomotion on Dynamic Rigid Surfaces} 

\author{Amir~Iqbal$^{1}$,~
        Sushant~Veer$^{2}$, and~
        Yan~Gu$^{1,\dagger}$
\thanks{$^{1}$A. Iqbal and Y. Gu are with the Department of Mechanical Engineering, University
of Massachusetts Lowell, Lowell, MA 01854, U.S.A.
{\tt\small amir\_iqbal@student.uml.edu, yan\_gu@uml.edu.}}
\thanks{$^{2}$S. Veer is with NVIDIA Research, Santa Clara, CA 95051, U.S.A. This work was conducted while S. Veer was with Princeton University.
{\tt\small sveer@nvidia.com.}}
\thanks{$^{\dagger}$Corresponding author: Y. Gu.}
}


\maketitle

\begin{abstract}

Legged robot locomotion on a dynamic rigid surface (i.e., a rigid surface moving in the inertial frame) involves complex full-order dynamics that is high-dimensional, nonlinear, and time-varying.
Towards deriving an analytically tractable dynamic model, this study theoretically extends the reduced-order linear inverted pendulum (LIP) model from legged locomotion on a stationary surface to locomotion on a dynamic rigid surface (DRS).
The resulting model is herein termed as DRS-LIP.
Furthermore, this study introduces an approximate analytical solution of the proposed DRS-LIP that is computationally efficient with high accuracy.
To illustrate the practical uses of the analytical results, they are used to develop a hierarchical planning framework that efficiently generates physically feasible trajectories for DRS locomotion.
The effectiveness of the proposed theoretical results and motion planner is demonstrated both through simulations and experimentally on a Laikago quadrupedal robot that walks on a rocking treadmill.

\end{abstract}


\IEEEpeerreviewmaketitle

\vspace{-0.15 in}
\section{Introduction}
\vspace{-0.05 in}
Legged robots have the potential to traverse various challenging surfaces, including stationary (uneven or discrete) surfaces~\cite{anymal_16,fawcett2021robust,zhang2021efficient, winkler2015planningSemini} and dynamic rigid surfaces (i.e., rigid surfaces that move in the inertial frame)~\cite{iqbal2020provably,gao2021invariant}.
Common real-world examples of the dynamic rigid surface (DRS) are ships, public transportation vehicles, trains, and elevators.
This study aims to model and analyze the essential dynamic behaviors of a legged robot that walks on a DRS by building a reduced-order model and deriving its analytic approximate solution, so as to provide physical insights into the robot dynamics as well as to produce analytic results that can be used for efficient planning of legged locomotion.
Yet, reduced-order modeling of DRS locomotion is fundamentally complex due to the nonlinear and high-dimensional nature of legged locomotion dynamics~\cite{motahar2016composing} and the time-varying movement of the surface-foot contact points~\cite{iqbal2020provably}.

\vspace{-0.15 in}
\subsection{Reduced-order models of stationary surface locomotion}
\vspace{-0.05 in}

A reduced-order dynamic model of legged locomotion captures the robot's essential dynamic behaviors.
One of the most widely studied reduced-order model for stationary surface locomotion is the linear inverted pendulum (LIP) model~\cite{kajita20013d}, which models a legged robot as a point mass atop a massless leg. 
Although the LIP does not capture the complete robot dynamics, 
many of today's legged robots can be relatively accurately described by the LIP, including bipeds~\cite{pratt2006capture,zhao2017robust} and quadrupeds~\cite{winkler2015planningSemini,mastalli2020motion}.
This is because they typically have a heavy upper body and lightweight legs.

Due to its simplicity, the LIP model is analytically tractable and can provide physical insights into the essential robot behaviors. 
It also explicitly reveals the relationship between the center of pressure (CoP), which can infer the feasibility of ground contact forces (i.e., whether the robot rolls about any edge of the region of contact), and the center of mass (CoM).
Thus, the LIP can be exploited to represent the robot dynamics in motion planning for ensuring the computational efficiency and physical feasibility of planning.
Beside motion planning, the LIP model has also been used as a basis to form balance~\cite{Kajita2003BipedWP} and stability~\cite{pratt2006capture} criteria, and its inherent connection with another widely studied reduced-order model, hybrid zero dynamics~\cite{westervelt2007feedback}, has recently been analyzed~\cite{gong2020angular}.

The LIP model has been extended to more complex scenarios by considering a varying CoM height~\cite{LIP_varyingHeight_caron2020biped}, CoM movement on a 3-D plane~\cite{zhao2017robust}, nontrivial centroidal angular momentum~\cite{pratt2006capture}, and hybrid robot dynamics~\cite{xiong2020global}.
Since these models are built upon the assumption that the foot-surface contact point is stationary, they may not be valid for legged locomotion on a DRS, especially when the surface motion is relatively significant.

\vspace{-0.15 in}
\subsection{Reduced-order models of dynamic surface locomotion}

For legged locomotion on a rigid surface whose motions are affected by the robot (e.g., passive and relatively lightweight surfaces), several reduced-order models have been derived and analyzed, including the LIP~\cite{nagarajan2014balancing,Yamenzheng2011ball}, centroidal dynamics~\cite{BallMan2020Koshil_yang}, and rimless-wheel model~\cite{asano2021modeling}.
Yet, these models may not be valid for robot walking over a rigid surface whose motion is not affected by the robot (e.g., trains, vessels, and elevators).
For these substantially heavy or rigidly actuated surfaces, the effects of the surface motion on a spring-loaded inverted pendulum has been numerically studied~\cite{iqbal_SLIP}.
Still, it is unclear under what conditions this model is valid for real-world DRS locomotion.

Beyond the scope of legged locomotion, the modeling and analysis of an inverted pendulum with a vertically oscillating support is a classical physics problem.
A well-known example is the Kapitza pendulum~\cite{kapitza1951dynamic}, which has an intriguing property that
under high-frequency support oscillation, the pendulum's upper equilibrium becomes stable whereas its lower equilibrium is unstable.
However, it remains unknown whether and when the Kapitza pendulum is a reasonable simplification of DRS locomotion.
Also,
in practical real-world applications, such as legged locomotion on a vessel, the surface motion frequency is typically too low to meet the conditions underlying the Kapitza pendulum.

\vspace{-0.15 in}
\subsection{Contributions}

This study aims to analytically extend the LIP model~\cite{kajita20013d} to substantially heavy or rigidly actuated DRSes (e.g., vessels and elevators) and to demonstrate the uses of the theoretical results in motion planning. 
The main contributions are:
\begin{enumerate}
    \item [(a)] Deriving a reduced-order model that describes the essential dynamic behaviors of a legged robot during DRS locomotion, by expanding the LIP model to DRS locomotion, which we term as ``DRS-LIP''.
    \item [(b)] Forming the analytical approximate solution of the DRS-LIP
    under a vertical, sinusoidal DRS motion, and giving physical insights into the model's stability.
    \item [(c)] Developing a hierarchical planner that utilizes the proposed DRS-LIP and its solution to efficiently generate feasible reference trajectories for DRS locomotion.
    \item [(d)] Assessing the accuracy and computational cost of the proposed solution with MATLAB simulations.
    \item [(e)] Validating the planner efficiency and feasibility through both physics-based simulations and hardware experiments, highlighting the practical usefulness of the proposed analytical results for motion planning.
\end{enumerate}
The derivation of the DRS-LIP model, i.e., item (a), has been partly reported in~\cite{iqbal2021extended}.
The new, substantial contributions are items (b)-(e), which are all previously missing.

The paper is organized as follows.
Section~\ref{sec: model} introduces the derivation of the proposed DRS-LIP model.
Section~\ref{sec: solution} presents the analytical approximate solution of the DRS-LIP model under a vertical, sinusoidal surface motion.
Section~\ref{sec: planning} develops an efficient motion planner based on the analytical results.
Section~\ref{sec: results} reports the validation results.
Section~\ref{sec: discussion} discusses the capabilities and limitations of the proposed methods.
Section~\ref{sec: conclusion} provides the conclusion remarks.

\vspace{-0.12 in}
\section{Reduced-Order Model of DRS Locomotion}
\vspace{-0.05 in}
\label{Modeling}
\label{sec: model}

This section introduces a reduced-order model that captures the essential dynamics of a legged robot that walks on a DRS.
The model is derived by extending the classical LIP model from static surfaces~\cite{kajita20013d} to a DRS.

Many of today's legged robots have a heavy upper body and lightweight legs.
Their center of mass (CoM) dynamics can be approximately described by an LIP, i.e., a point mass atop a massless leg~\cite{kajita20013d}, under the assumption that:
\begin{itemize} 
    \item [(A1)] The robot's rate of whole-body angular momentum about the CoM is negligible.
\end{itemize}
Assumption (A1) is reasonable for real-world robot locomotion because the trunk is typically controlled to maintain a steady posture for housing sensors (e.g., cameras).

\begin{figure}[t]
    \centering
    \includegraphics[width= 0.9\linewidth]{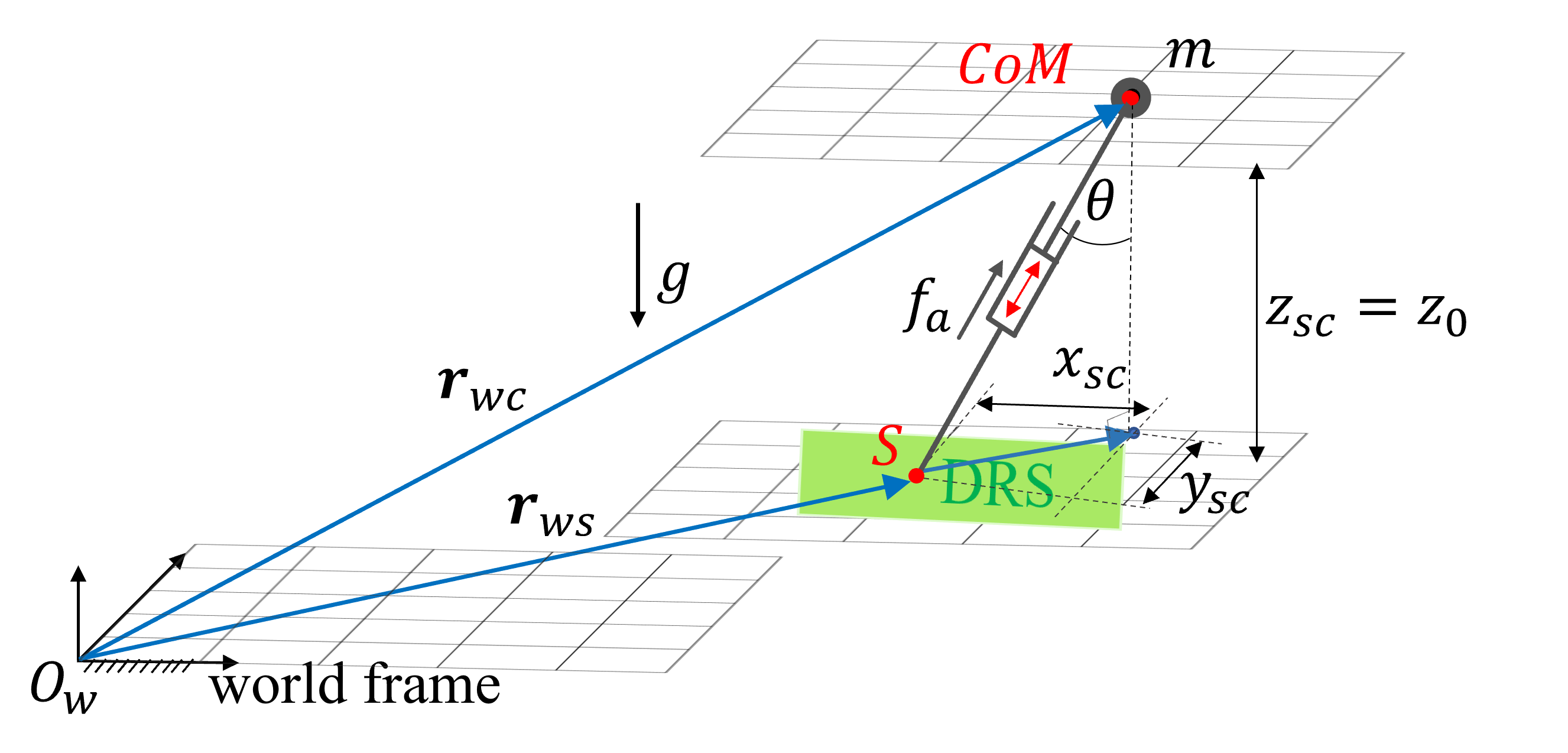}
    \vspace{-0.1 in}
    \caption{Illustration of the proposed DRS-LIP model.
    All three grid planes are horizontal. 
    The top and middle ones pass through the CoM and the leg's far end $S$, respectively.
    The bottom one is fixed to the world frame.}
    \label{Fig:LIPM_sketch}
        \vspace{-0.2 in}
\end{figure}

In this study, we use a three-dimensional (3-D) LIP to capture the essential dynamics of a 3-D legged robot walking on a DRS (see Fig.~\ref{Fig:LIPM_sketch}).
The point mass and support point {\small $S$} in Fig.~\ref{Fig:LIPM_sketch} correspond to the robot's CoM and CoP.

Let {\small $\mathbf{r}_{wc}=[x_{wc},~y_{wc},~z_{wc}]^T$} and {\small $\mathbf{r}_{ws}=[x_{ws},~y_{ws},~z_{ws}]^T$} respectively denote the position of the CoM and point {\small $S$} in the world frame.
Then, the CoM position relative to point $S$, denoted as {\small $\mathbf{r}_{sc}$}, is defined as: {\small $\mathbf{r}_{sc}=\mathbf{r}_{wc} -\mathbf{r}_{ws}=[x_{sc},~y_{sc},~z_{sc}]^T$}.

The CoM dynamics during DRS locomotion is given by:
\vspace{-0.05 in}
\begin{equation}
\small
    \ddot{x}_{wc} = \frac{f_a x_{sc}}{m r} \sin \theta,
    ~ \ddot{y}_{wc} = \frac{f_a y_{sc}}{m r} \sin \theta ,
    ~
    \ddot{z}_{wc} = \frac{f_a}{m} \cos \theta -g.
     \label{Eq: LIPM_EOM_world - z}
     \vspace{-0.05 in}
\end{equation}
Here, {\small $m$} is the total mass,
{\small $\theta$} is the angle of the vector {\small $\mathbf{r}_{sc}$} relative to the vertical axis,
{\small $g$} is the norm of the gravitational acceleration vector,
and {\small $r$} is the projected length of {\small $\mathbf{r}_{sc}$} on the horizontal plane.
The scalar variable {\small $f_a$} is the norm of the ground contact force. 

\vspace{-0.15 in}
\subsection{DRS-LIP under a General Vertical Surface Motion}

We consider the following assumption on the vertical distance {\small $z_{sc}$} between the CoM and point {\small $S$} (see Fig.~\ref{Fig:LIPM_sketch}):
\begin{itemize} 
    \item[(A2)] The CoM maintains a constant height $z_0$ above the support point {\small $S$} (i.e., {\small $z_{sc} = z_0$}).
\end{itemize}
This assumption is analogous to the simplifying assumption of the classical LIPM model that the point-mass height over the stationary surface is constant~\cite{kajita20013d}.

Under assumption (A2), the relationships
{\small $\dot{z}_{wc} = \dot{z}_{ws}$} and {\small $\ddot{z}_{wc} = \ddot{z}_{ws}$} hold, and then the axial force {\small $f_a$} becomes
{\small $ f_a = m (\ddot{z}_{ws}+g)/\cos \theta$}.
Thus, the horizontal LIP dynamics is:
\vspace{-0.05 in}
\begin{equation}
\small
     \ddot{x}_{wc} =
     (\ddot{z}_{ws}+g)\frac{x_{sc}}{z_0}
     ~\text{and}~
    \ddot{y}_{wc}= 
    (\ddot{z}_{ws}+g)\frac{y_{sc}}{z_0}.
     \label{Eq:XY-motion general}
     \vspace{-0.05 in}
\end{equation}
Then, by substituting {\small $\ddot{x}_{wc}=\ddot{x}_{ws}+ \ddot{x}_{sc}$} and {\small $\ddot{y}_{wc}=\ddot{y}_{ws}+ \ddot{y}_{sc}$} into, \eqref{Eq:XY-motion general}, the horizontal LIP dynamics becomes:
\vspace{-0.05 in}
\begin{equation}
\small
    \ddot{x}_{sc}  - \frac{(\ddot{z}_{ws}+g)}{z_0} x_{sc} = -\ddot{x}_{ws}
   ~ \text{and}~
    \ddot{y}_{sc}  - \frac{(\ddot{z}_{ws}+g)}{z_0} y_{sc} = -\ddot{y}_{ws}.
    \label{Eq:simplified_Cap-xy}
    \vspace{-0.05 in}
\end{equation}
When there is no slippage between the support point {\small $S$} and the surface, the acceleration of point {\small $S$}, i.e., {\small $(\ddot{x}_{ws},\ddot{y}_{ws},\ddot{z}_{ws})$}, equals the DRS's acceleration at that point.
Given that real-world DRSes (e.g., vessels) are typically equipped with high-accuracy, real-time motion monitoring systems~\cite{Ship_Motion_Monitoring_System}, we assume the time profile of {\small $(\ddot{x}_{sw},\ddot{y}_{sw},\ddot{z}_{sw})$} is known.
Accordingly, they are treated as explicit time functions.
Thus, the dynamics in~\eqref{Eq:simplified_Cap-xy} is linear, nonhomogenous, and time-varying.

Since DRSes, such as cruising ships in regular sea waves, have relatively small horizontal acceleration compared with vertical acceleration~\cite{gahlinger2000const_hv,Vessels_Vertical_Periodic_1954},
we assume that the horizontal acceleration of point {\small $S$} is sufficiently small to be ignored:
\begin{itemize} 
    \item[(A3)] The horizontal acceleration of point {\small $S$} (i.e., {\small $\ddot{x}_{ws}$} and {\small $\ddot{y}_{ws}$}) are negligible.
\end{itemize}
Then, the forcing terms in \eqref{Eq:simplified_Cap-xy} (i.e., {\small $-\ddot{x}_{ws}$} and {\small $-\ddot{y}_{ws}$}) can be approximated as zero, and the horizontal LIP dynamics in \eqref{Eq:simplified_Cap-xy} becomes linear, time-varying, and homogeneous:
\vspace{-0.05 in}
\begin{equation}
\small
    \ddot{x}_{sc}  - \frac{(\ddot{z}_{ws}+g)}{z_0} x_{sc} = 0
    ~\text{and}~
    \ddot{y}_{sc}  - \frac{(\ddot{z}_{ws}+g)}{z_0} y_{sc} = 0.
    \label{Eq-LIPM_on_DRS_simplified} 
\vspace{-0.05 in}
\end{equation}
Note that the vertical CoM trajectory is given by {\small $z_{sc} = z_0$}.

\noindent \textbf{Remark 1:}
The LIP in \eqref{Eq-LIPM_on_DRS_simplified}, along with {\small $z_{sc} = z_0$}, describes the simplified dynamics of a robot walking on a DRS under assumptions (A1)-(A3), which we call ``DRS-LIP''.

\vspace{-0.15 in}
\subsection{DRS-LIP under a Vertical Sinusoidal Surface Motion}

A real-world DRS, such as a vessel in regular sea waves, typically exhibits a vertical, sinusoidal motion with a constant amplitude and frequency~\cite{Vessels_Vertical_Periodic_1954}.
Thus, we focus on such motions for further analystis of the DRS-LIP. 

Under a vertical, sinusoidal surface motion, the vertical acceleration {\small $\ddot{z}_{ws}$} of point {\small $S$} is sinusoidal, and \eqref{Eq-LIPM_on_DRS_simplified} becomes the well-known Mathieu's equation~\cite{farkas2013periodic}.


Without loss of generality, the vertical sinusoidal motion of the DRS at the surface-foot contact point is assumed as:
\vspace{-0.05 in}
\begin{equation}
\small
    z_{ws} = A \sin \omega t,
    \label{Eq-ref: sinusoidal DRS motion}
    \vspace{-0.05 in}
\end{equation}
where the real, scalar parameters {\small $A$} and {\small $\omega$} are the amplitude and frequency of the vertical motion, respectively.

Therefore, the surface acceleration {\small $\ddot{z}_{ws}$} at the support point is expressed as: {\small $\ddot{z}_{ws} :=-A \omega^2 \sin \omega t$}.
Substituting the expression of {\small $\ddot{z}_{ws}$}
in \eqref{Eq-LIPM_on_DRS_simplified}, we have:
    \vspace{-0.05 in}
\begin{equation}
        \ddot{x}_{sc}  - \tfrac{(g-A \omega^2 \sin \omega t)}{z_0} x_{sc} = 0
        ~ \text{and}~         
        \ddot{y}_{sc}  - \tfrac{(g-A \omega^2 \sin \omega t)}{z_0} y_{sc} = 0.
        \label{Eq-Hills_vs_motion}
            \vspace{-0.05 in}
\end{equation}
As the equations in the {\small $x$}- and {\small $y$}-directions are decoupled and share the same structure, their solutions share the same form.
Thus, for brevity, we focus on deriving the solution along the $x$-direction, {\small ${x}_{sc}$}, in the next section.

The DRS-LIP in~\eqref{Eq-Hills_vs_motion} can be transformed into the standard
Mathieu's equation~\cite{farkas2013periodic} by introducing a new time variable $\tau :=\frac{{\pi}+2\omega t}{4}$ and rewriting \eqref{Eq-Hills_vs_motion} in terms of $\tau$ as:
    \vspace{-0.05 in}
\begin{equation}
\small
    \frac{d^2x_{sc}}{d\tau^2} +(c_0 - 2c_1\cos 2 \tau)x_{sc}  = 0,
    \label{Eq-transformed_MathieuEqn}
        \vspace{-0.05 in}
\end{equation}
where the real scalar coefficients $c_0$ and $c_1$ are defined as $c_0 := -\frac{4g}{\omega^2 z_0}$ and $c_1 := \frac{2A}{z_0}$.

\vspace{-0.12 in}
\section{APPROXIMATE ANALYTICAL SOLUTION}
\vspace{-0.05 in}
\label{sec: solution}

This section introduces an approximate analytical solution of the DRS-LIP under a vertical, sinusoidal DRS motion.

\vspace{-0.18 in}
\subsection{Computation of Approximate Analytical Solutions}

The DRS-LIP model in~\eqref{Eq-Hills_vs_motion} generally does not have an exact, closed-form analytical solution.
One approach to derive an approximate analytical solution is to utilize the fundamental solution matrix based on the Floquet theory~\cite{farkas2013periodic}.
Alternatively, we choose to exploit the existing analytical results of the well-studied Mathieu's equation to obtain a more computationally efficient solution.

There are various existing analytical approximate solutions 
of Mathieu's equation, including periodic solutions~\cite{phelps1965analytical} and those expressed through power series~\cite{farkas2013periodic}.
In this study, we adopt the general, exact analytical solution from~\cite{BookB_ME_werth2005charged} because of its generality and computationally efficiency:

\noindent \textbf{Theorem:}
The exact, general (periodic or non-periodic) analytical solution of the DRS-LIP in~\eqref{Eq-transformed_MathieuEqn} takes the following form:
\vspace{-0.05 in}
\begin{equation}
\small
    {x}_{sc}(\tau) = \alpha_1 e^{\mu \tau}\sum_{n=-\infty}^{\infty}C_{2n} e^{i2n\tau} +\alpha_2 e^{-\mu \tau}\sum_{n=-\infty}^{\infty}C_{2n} e^{-i2n\tau}.
    \label{Eq-Assumed_GenSol_of_MathieuEqn}
\vspace{-0.05 in}
\end{equation}
Here, {\small $\mu$} is the characteristic exponent of~\eqref{Eq-transformed_MathieuEqn}.
The coefficients {\small $\alpha_1$} and {\small $\alpha_2$} are real scalars, the scalar {\small $n$} is an integer, and the coefficients {\small $C_{2n}$}'s are complex scalars.
\hfill $\blacksquare$

The proof of Theorem 1 can be readily obtained based on~\cite{BookB_ME_werth2005charged}.
To use~\eqref{Eq-Assumed_GenSol_of_MathieuEqn} to compute an approximate solution,
we need to determine the number of terms to keep in the approximate solution as well as the values of the parameters {\small $\mu$}, {\small $\alpha_1$}, {\small $\alpha_2$}, and {\small $C_{2n}$}'s~\cite{BookB_ME_werth2005charged}, which is explained next.

\subsubsection{Computing characteristic exponent {\small $\mu$}}
Substituting the solution \eqref{Eq-Assumed_GenSol_of_MathieuEqn} into \eqref{Eq-transformed_MathieuEqn} yields a recurrence relationship:
\vspace{-0.05in}
\begin{equation}
\small
    \beta_n(\mu)C_{2(n-1)} + C_{2n} + \beta_n(\mu)C_{2(n+1)} =0,
    \label{Eq: recurrence relation}
\vspace{-0.05in}
\end{equation}
where the scalar, complex coefficient function {\small $\beta_n$} is given by {\small $\beta_n(\mu):=\frac{c_1}{(2n-i\mu)^2-c_0}.$}
The derivation of \eqref{Eq: recurrence relation} is given in~\cite{BookB_ME_werth2005charged} and the supplementary material.

Equation \eqref{Eq: recurrence relation} for all {\small $n\in \mathbb{Z}^+$}
generates the following infinite set of linear homogeneous equations with the coefficients {\small $C_{2n}$}'s as the unknown variables:
{\small $
\boldsymbol{\Delta}(\mu)
\begin{bmatrix}
 \cdots,~C_{-6},~C_{-4},~C_{-2}~C_0,~C_2,~C_4,~C_6,~\cdots
 \end{bmatrix}^T
  =
 \mathbf{0},$}
where {\small $\mathbf{0}$} is an infinity-dimensional zero column vector and
\vspace{-0.05 in}
\begin{equation}
\small
    \boldsymbol{\Delta}(\mu)
:=
    \begin{bmatrix}
\ddots  & \vdots & \vdots & \vdots & \vdots & \vdots & \vdots & \vdots & \reflectbox{$\ddots$} \\ 
\cdots &0 & \beta_{-1} & 1 &\beta_{-1} &0 &0 &0 &\cdots \\
\cdots &0 &0 & \beta_0 & 1 &\beta_0 &0 &0 &\cdots \\
\cdots  &0 &0 &0 & \beta_1 & 1 &\beta_1 &0 &\cdots \\
\reflectbox{$\ddots$}  & \vdots & \vdots & \vdots & \vdots & \vdots  & \vdots & \vdots & \ddots \\ 
 \end{bmatrix}.
 \vspace{-0.05 in}
\end{equation}

This set of linear equations have
nontrivial solutions for the coefficients {\small $C_{2n}$}'s if the determinant of {\small $\boldsymbol{\Delta}(\mu)$}, denoted as {\small $ \text{det}( \boldsymbol{\Delta}(\mu) )$}, equals zero.
From \cite{Hills_det_Simplification_bateman1953higher}, we know the equation $\text{det}(\boldsymbol{\Delta} (\mu)  ) =0$ can be simplified as:
{\small $    2 |\boldsymbol{\Delta}(0)| \sin^2{\tfrac{\pi \sqrt{c_0}}{2}}  = 1- \cosh {\mu \pi}.
    \label{Eq: Characteristic exponent_Simplified}$}
Thus, we can obtain {\small $\mu$} through:
\vspace{-0.05in}
\begin{equation}
\small
\mu =\frac{1}{\pi} \cosh^{-1}({1-2  |\boldsymbol{\Delta}(0)|  \sin^2{\frac{\pi \sqrt{c_0}}{2}}}).
\label{Eq: mu}
\vspace{-0.05in}
\end{equation}

\noindent \textbf{Remark 2:}
Given the coefficient {\small $c_0$} of a DRS-LIP model, we can pre-compute {\small $\mu$} using~\eqref{Eq: mu}, \eqref{Eq: mu}, and the expressions of {\small $\boldsymbol{\Delta}(\mu)$}, which could then be used to compute the analytical solution during online planning.

\subsubsection{Determining {number of terms kept}}

The infinite series in the exact solution \eqref{Eq-Assumed_GenSol_of_MathieuEqn} is convergent because {\small $ \displaystyle{\lim_{n\to\infty} C_{2n} =0.} $}
Specifically,
{\small $C_{2n}$} is proportional to {\small $\beta_n(\mu)$}
and
{\small $ \displaystyle{\lim_{n\to\infty}  \beta_n(\mu) =0}$} (see supplementary material for proof).
Thus, we can approximate the exact solution with the sum of finite terms.

With {\small $N$} terms kept,
the approximate 
solution is given by:
\vspace{-0.05 in}
\begin{equation}
\small
    \hat{x}_{sc}(\tau) = \alpha_1 e^{\mu \tau}
    {\sum_{n=-N}^{N}} C_{2n} e^{i2n\tau} +\alpha_2 e^{-\mu \tau}\sum_{n=-N}^{N}C_{2n} e^{-i2n\tau}.
    \label{Eq-Assumed_GenSol_of_MathieuEqn_approx}
\vspace{-0.05 in}
\end{equation}

To ensure a sufficient solution accuracy, we can determine $N$ by numerically evaluating its effect on the accuracy.

\subsubsection{Computing {coefficients} {\small $C_{2n}$}, {\small $\alpha_1$}, and {\small $\alpha_2$} }

With {\small $\mu$} computed, we can determine the value of the coefficients {\small $C_{2n}$} recursively from \eqref{Eq: recurrence relation} by setting {\small $C_{2N}=0$} and {\small $C_0=A$}~\cite{BookB_ME_werth2005charged}.
The values of {\small $\alpha_1$ and $\alpha_2$} can be obtained based on the initial conditions {\small $\hat{x}_{sc}(0)$} and {\small $\dot{\hat{x}}_{sc}(0)$}.
Details of the computations are given in supplementary material.


The last step of determining the approximate analytical solution is to transform the solution {\small $\hat{x}_{sc}(\tau)$} into the usual time domain {\small $t$} using {\small $\tau=({\pi}/{2}+\omega t)/2$}.

\vspace{-0.15 in}
\subsection{Stability Analysis}
\label{subsec: Stability}
\vspace{-0.05 in}

By the Floquet theory, the DRS-LIP model in~\eqref{Eq-Hills_vs_motion} is called ``stable'' if all its solutions are bounded for all {\small $t>0$}, and is ``unstable'' if an unbounded solution exists for {\small $t>0$}.
The stability properties of the DRS-LIP model can be determined with the characteristic exponents {\small $\mu$}.
Since the DRS-LIP is a linear, second-order ordinary differential equation, it has two characteristic exponents, denoted as {\small $\mu_1$} and {\small $\mu_2$}.
By the Floquet theory, the model is stable if and only if {\small $\text{Re}(\mu_1),\text{Re}(\mu_2) <0$}.
Suppose that {\small $\text{Re}(\mu_1) < \text{Re}(\mu_2)$}.
Validation of this physical insight is given in Sec.~\ref{sec: results}.

\vspace{-0.1 in}
\section{DRS-LIP BASED HIERARCHICAL PLANNING}

\label{sec: planning}

To illustrate the practical uses of the DRS-LIP model and its approximate analytical solution, this section presents a hierarchical planner that exploits them to enable efficient and feasible planning of quadrupedal walking on a DRS.

The planner is designed for quadrupedal walking~\cite{iqbal2020provably,mastalli2020motion} whose complete gait cycle comprises four continuous foot-swinging phases and four discrete foot-landing events, as illustrated in Fig.~\ref{Fig:Walking_phases}.
This planner also assumes the DRS motion is known, which is realistic for real-world DRSes such as vessels because they typically possess motion monitoring systems~\cite{ShipMotion_tannuri2003estimating}.

\begin{figure}[t]
    \centering
    \includegraphics[width=0.95\linewidth]{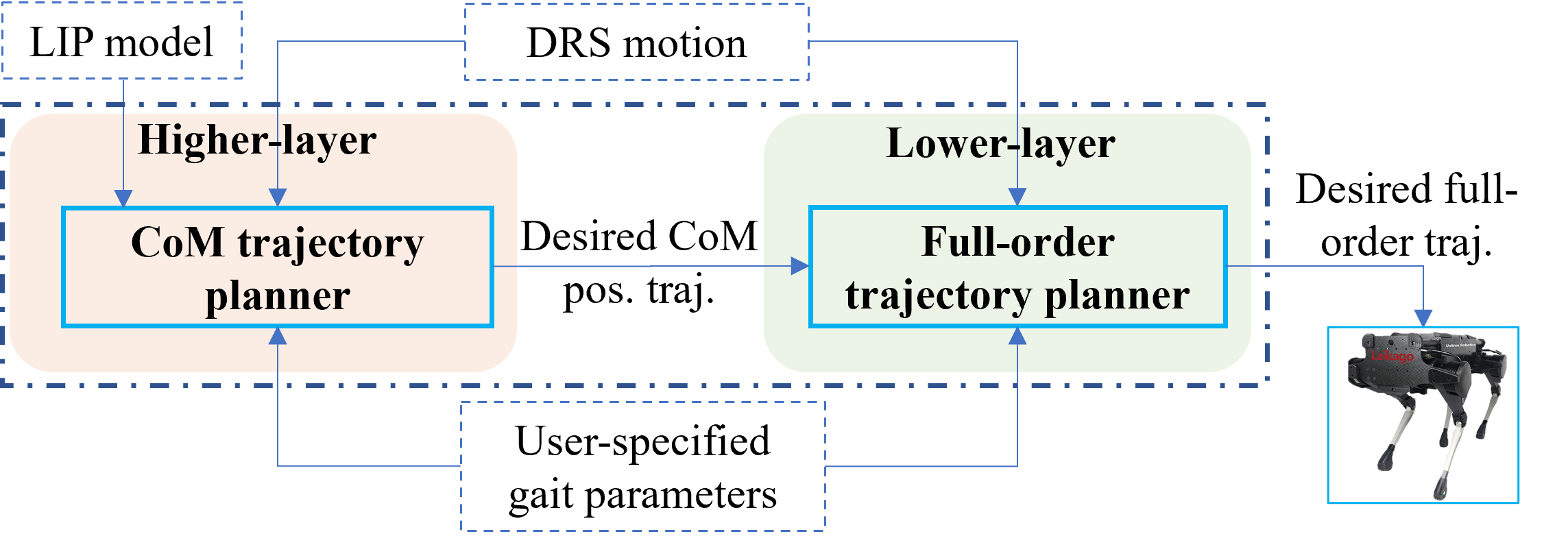}
        \vspace{-0.1 in}
    \caption{Overview of the proposed hierarchical planner. \YanCom{  }}
    \label{Fig:framework}
    \vspace{-0.2 in}
\end{figure}

The planner has two layers (see Fig.~\ref{Fig:framework}).
The higher layer produces kinematically and dynamically feasible CoM trajectories for the DRS-LIP model of a legged robot by incorporating necessary feasibility constraints.
The lower layer uses trajectory interpolation to translate the CoM trajectories
into the desired motion for all degrees of freedom of the full-order robot model.

\vspace{-0.15 in}

    \vspace{-0.15 in}
\subsection{Higher-Layer CoM Trajectory Planner}

The higher planner uses the DRS-LIP model to efficiently generate feasible reference CoM position trajectories for the complete gait cycle, by solving the following nonlinear optimization problem. 

\subsubsection{User-defined gait parameters}
The input to the higher-layer planner is the user-defined gait parameters (which specify the desired gait features) and the known DRS motion (which is vertical and sinusoidal).
The gait parameters are commonly chosen as:
(i) average walking velocity (i.e., horizontal CoM velocity),
(ii) foot-contact sequence,
(iii) stance-foot positions,
(iv) constant CoM height above the surface (for respecting assumption (A2)) and
(v) gait period.
The values of parameters (i)-(iv) are typically set to help ensure a kinematically feasible gait.
The value of parameter (v) is selected such that the quotient of the DRS's motion period and the desired gait period is an integer (i.e., the desired CoM motion complies with the DRS motion).

\subsubsection{Optimization variables}
We choose the optimization variables {\small $\boldsymbol{\alpha}$} of the planner as the
initial horizontal CoM position
and velocity
of each continuous phase.
The rationale for this choice is that given the DRS-LIP model parameters these variables completely determine
the horizontal CoM trajectories.
The vertical CoM position is not included as an optimization variable because it can be readily obtained from the known DRS motion and the user-defined CoM height.

\subsubsection{Constraints}
We choose to design the constraints to help enforce gait feasibility and to respect the desired gait features as specified by the user-defined parameters. 
Note that these constraints are formed based on the proposed analytical approximate solution.
The equality constraints include:
(i) continuity of the CoM trajectories at the foot-landing events
and
(ii) the user-specified walking velocity.
The inequality constraints are:
(i) friction cone constraint for avoiding foot slipping,
(ii) confinement of CoM trajectories within the polygon of support
for approximately respecting the CoP constraint, and
(iii) upper and lower bounds on the optimization variables {\small $\boldsymbol{\alpha}$}.

\begin{figure}[t]
    \centering
    \includegraphics[width=0.85\linewidth]{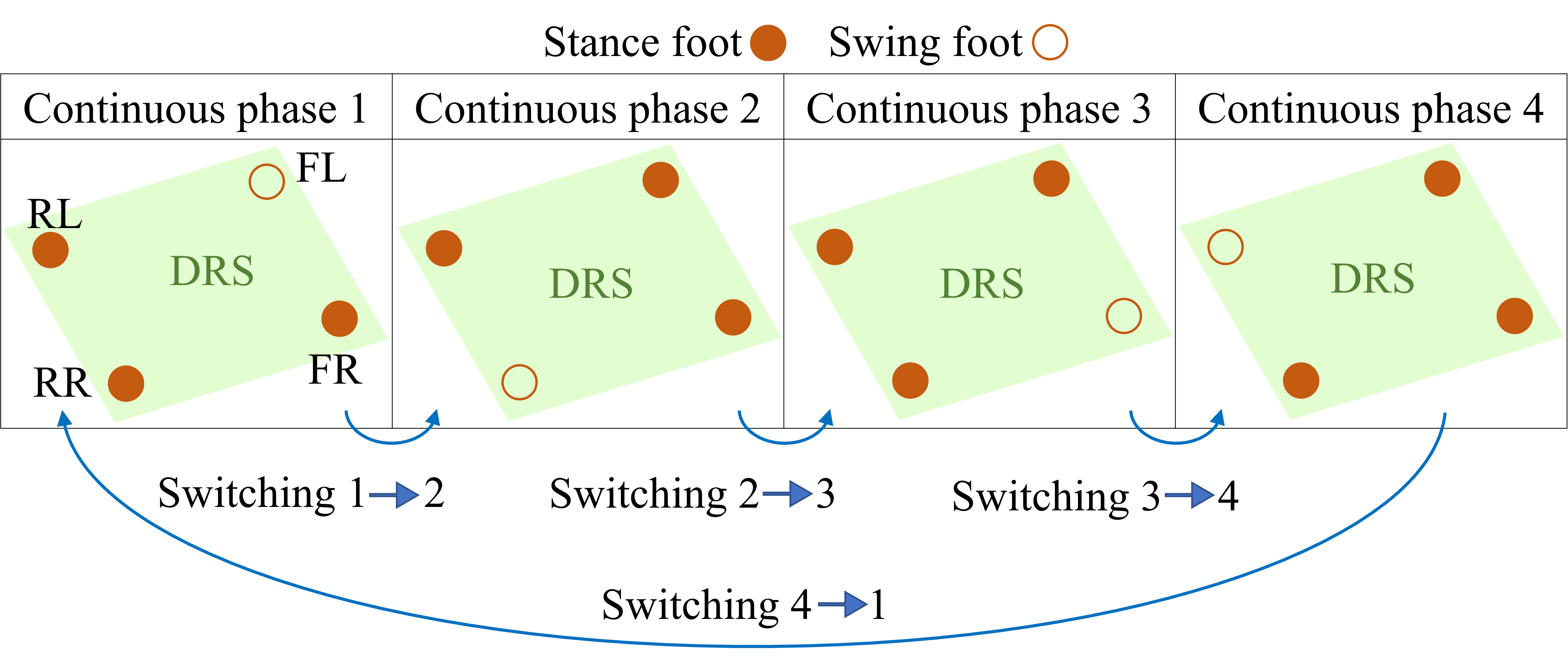}
    \vspace{-0.1 in}
    \caption{ A complete cycle of quadrupedal walking on a DRS.}
    \label{Fig:Walking_phases}
    \vspace{-0.2 in}
\end{figure}

To meet the constraints, {\small $\boldsymbol{\alpha}$} is solved through:
\vspace{-0.05 in}
\begin{equation}
\small
\begin{aligned}
\small
\min_{\boldsymbol{\alpha}} \quad &  \small h(\boldsymbol{\alpha})
\\
\small
\textrm{subject to} \quad &  \mathbf{f}_{eq}(\boldsymbol{\alpha}) =\mathbf{0},
~
\mathbf{g}_{ineq}(\boldsymbol{\alpha}) \leq \mathbf{0},
  \label{Optimization_P1}
\end{aligned}
\vspace{-0.05 in}
\end{equation}
where {\small $h(\boldsymbol{\alpha})$} is a scalar cost function (e.g., energy cost of transport), and the vector functions {\small $\mathbf{f}_{eq}$} and {\small $\mathbf{g}_{ineq}$} are the sets of all equality and inequality constraints, respectively.

\vspace{-0.15 in}
\subsection{Lower-Layer Full-Body Trajectory Planner}
\vspace{-0.05 in}

The lower-layer planning is essentially trajectory interpolation that translates the reference CoM position trajectory supplied by the higher-layer planner into the full-order trajectories of a quadrupedal robot in Cartesian space.
To impose a steady trunk/base pose and to avoid swing foot scuffing on the surface, we choose these full-order trajectories to be the absolute base pose and relative swing-foot position with respect to the base.

The input to the lower-layer planner (Fig. \ref{Fig:framework}) is: the known DRS motion that is vertical and sinusoidal, the  CoM trajectories provided by the higher-layer planner, and user-defined parameters (including robot's CoM height above the CoP, stance foot locations, and maximum swing foot height).

Furthermore, as inspired by previous quadrupedal robot planning~\cite{winkler2015planningSemini},
a brief four-leg-in-support phase is inserted upon a foot-landing event when the two consecutive polygons of support only share a common edge (i.e., ``Switching {\small $1\rightarrow 2$}'' and ``Switching {\small $3\rightarrow 4$}'' in Fig~\ref{Fig:Walking_phases}, so as to ensure smooth and feasible transitions during these events.

\subsubsection{Base trajectories}

The CoM of the robot is approximated as the base (i.e., the geometric center of the trunk) because a quadruped's trunk typically has a symmetric mass distribution and is substantially heavier than the legs.
Thus, the desired base position trajectories are provided by the higher-level planner.
To avoid overly stretched leg joints for ensuring kinematic feasibility, the desired base orientation trajectories are designed to comply with the DRS orientation.

\subsubsection{Swing foot trajectories}
The swing foot trajectories are designed to agree with the user-defined stance-foot locations and to respect the kinematic limits of the robot's leg joints.
Specifically, we obtain the desired swing foot trajectory during a continuous phase by using B\'ezier polynomials~\cite{iqbal2020provably} to connect the adjacent desired stance-foot positions.

\noindent \textbf{Remark 3:}
The dynamic feasibility of the planned trajectories depends on the closeness of the DRS-LIP and the actual robot dynamics.
The DRS-LIP model is a relatively faithful representation of an actual DRS-robot system when the system behavior reasonably respects assumptions underlying the proposed model and its solution.
Indeed, assumption (A3) holds when the known surface motion is vertical and sinusoidal,
and the planner explicitly imposes assumption (A2).
Moreover, although the planner enforces the desired base orientation to comply with the surface orientation for kinematic feasibility, assumption (A1) is still reasonably respected by the planned motion.
This is because the rate of the robot's centroidal angular momentum is negligible compared with that of its linear momentum under the typical angular movement range of real-world DRSes (e.g., vessels~\cite{ShipMotion_tannuri2003estimating}).

\vspace{-0.1 in}
\section{SIMULATION AND EXPERIMENT RESULTS}
\label{sec: results}

This section presents the validation results for the DRS-LIP model, analytical solution, and hierarchical planner.

\vspace{-0.15 in}
\subsection{Solution Validation}

\subsubsection{Validation of solution accuracy and efficiency} 

The accuracy and computational efficiency of the proposed analytical approximate solution in \eqref{Eq-Assumed_GenSol_of_MathieuEqn_approx} is assessed through comparison with the highly accurate numerical solution.
For fairness of comparison, both solutions are computed in MATLAB for the interval {\small $t \in [0, ~0.5]$} sec. 
The approximate solution has ten terms kept (i.e., {\small $N=10$}) for a reasonable trade-off between accuracy and computational efficiency.
The comparative numerical solution is computed using MATLAB's ODE45 solver with an error tolerance of {\small $10^{-9}$} and at {\small 1000} evenly distributed instants within the given interval.

To validate the proposed solution under different initial conditions, {\small 1000} sets of initial conditions are randomly chosen within {\small $|x_{sc}(0)|<0.2$} m and {\small $|\dot{x}_{sc}(0)|<0.2$} m/s.
The DRS-LIP model parameters are chosen to be within realistic ranges of DRS motions~\cite{Vessels_Vertical_Periodic_1954,ShipMotion_tannuri2003estimating} and quadrupedal robot dimensions: {\small $A =7$} cm, {\small $\omega = \pi$} rad/s, and {\small $z_0 =42$} cm.

\begin{figure}[t]
    \centering
    \includegraphics[width= 0.85\linewidth]{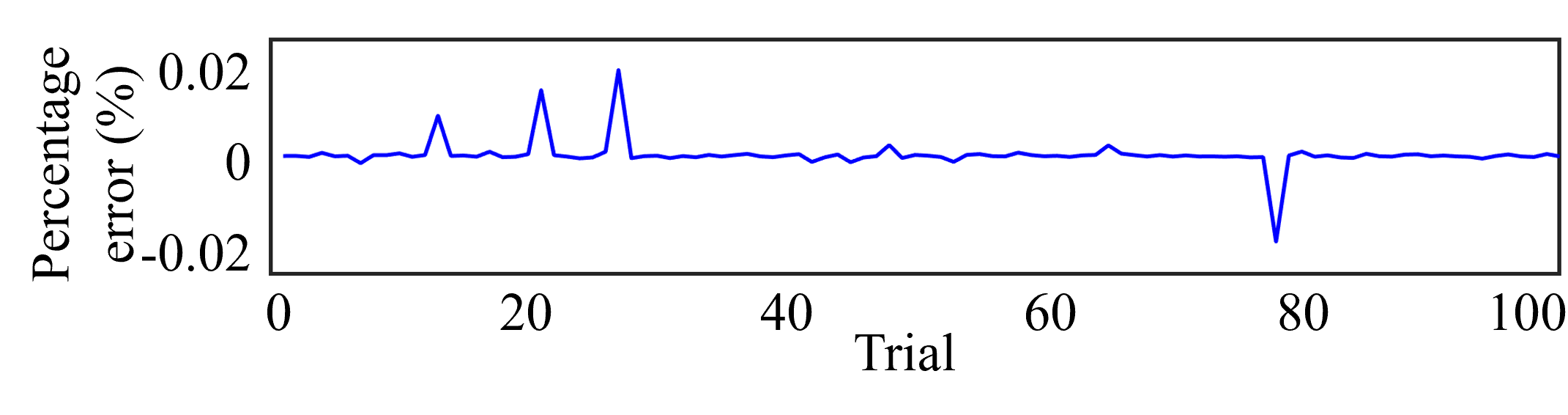}
    \vspace{-0.2 in}
    \caption{Mean percentage error of the proposed analytical approximate solution compared with high-accuracy numerical solution under model parameters {$A =7$} cm, {$\omega = \pi$} rad/s, and {$z_0 =42$} cm for 100 random initial conditions satisfying {$|x_{sc}(0)|<0.2$} m and {$|\dot{x}_{sc}(0)|<0.2$} m/s.}
    \label{Fig-SolAccuracy}
    \vspace{-0.2 in}
\end{figure}

Figure~\ref{Fig-SolAccuracy} shows the accuracy of the approximate analytical solution (with ten terms kept) compared with the numerical solution for {\small 100} out of the {\small 1000} trials.
Within the shown {\small 100} trials, the maximum value of the absolute percentage error is lower than {\small 0.02$\%$} in magnitude, indicating the reasonable accuracy of the proposed approximate solution.
The absolute percentage error, measured by mean $\pm$ one standard deviation, is {\small $(0.0012 \pm 0.005)\%$} for all {\small $1000$} trials.

\begin{figure}[t]
    \centering
    \includegraphics[width= 0.85\linewidth]{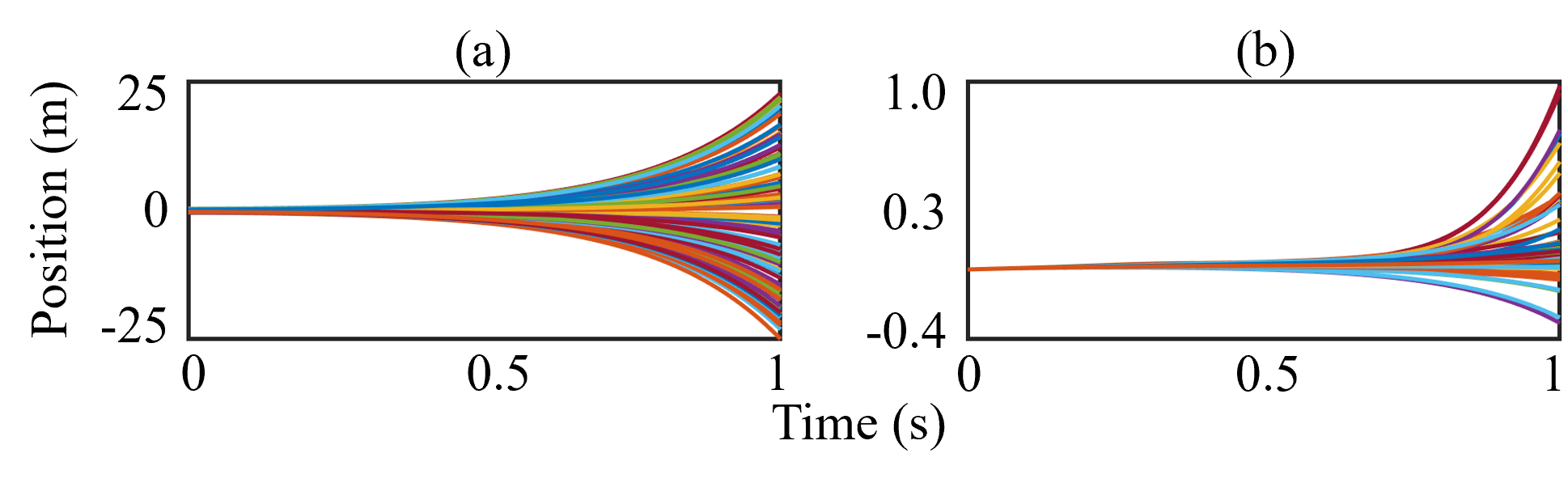}
    \vspace{-0.2 in}
    \caption{
    Unbounded time evolution of solution $\hat{x}_{sc}(t)$ under:
    (a) the same model parameters {$A =7$} cm, {$\omega = \pi$} rad/s, and {$z_0 =42$} cm but 100 different initial conditions satisfying $|x_{sc}(0)|<0.4$ m and $|\dot{x}_{sc}(0)|<0.4$ m/s
    and
    (b) different parameters ($0 <\omega\leq 2\pi$ rad/s, $0<A\leq 100$ cm, and $30\leq z_0 \leq 55$ cm) but the same initial condition $x_{sc}(0)=0.02$ m and $\dot{x}_{sc}(0)=0.1$ m/s.}
    \vspace{-0.2 in}
    \label{Fig-Sol_stability}
\end{figure}

Table \ref{table:Comparision_time_Sol} displays the comparison of the average computational time cost (measured by mean$\pm$SD) for {\small 1000} trials.
The approximate analytical solution is around {\small $15$} times faster to compute than the numerical one.

\begin{table}[h!]
\centering
\vspace{-0. in}
\caption{Average computation time of analytical and numerical solutions for 1000 trials in MATLAB (mean $\pm$ SD)}
\label{table:Comparision_time_Sol}
\vspace{-0.15 in}
\begin{tabular}{c| c c } 
 \hline
  \hline
 \small  Solution method & \small  Computation time (ms) 
\\[0.5ex]
 \hline\hline 
  \small Numerical  & \small  $2.61\pm 0.43$  \\[0ex]  
  \small Analytical & \small  $0.16\pm 0.02$  \\ [0ex] 
 \hline
\end{tabular}
\vspace{-0.1 in}
\end{table}

\subsubsection{Validation of stability property}

For typical ship motions in regular sea waves~\cite{ShipMotion_tannuri2003estimating}, the DRS-LIP model parameters take values as: {\small $A \leq 100$} cm and {\small $\omega \leq 2\pi$} rad/s. 
Also, the kinematically feasible CoM height {\small $z_0$} of a typical quadrupedal robot (e.g., Unitree's Laikago) is within {\small $[0.3, 0.55]$} m.
Under these parameter ranges, we use~\eqref{Eq: mu} to numerically compute the characteristic exponents and obtain that {\small $\text{Re}(\mu_2)>0$} and {\small $\text{Re}(\mu_1) <0$}.
Thus, by the Floquet theory, the DRS-LIP is unstable (i.e., an unbounded solution exists) under the considered operating condition.

Figure~\ref{Fig-Sol_stability} presents the validation results of this physical insight.
Subplot (a) displays the analytical approximate solutions under different initial conditions ({\small $|x_{sc}(0)|<0.4$} m and {\small $|\dot{x}_{sc}(0)|<0.4$} m/s) and DRS-LIP parameters ({\small $\omega=\pi$}  rad/s, $A=7$ cm, and {\small $z_0=42$ cm}).
Subplot (b) shows the solutions under the same initial condition ({\small $x_{sc}(0) =0.02$} m and {\small $\dot{x}_{sc}(0) =0.10$} m/s) but different model parameters ({\small $0 <\omega\leq 2\pi$} rad/s, {\small $0<A\leq 100$} cm, and {\small $30\leq z_0 \leq 55$} cm).
In all cases except for the trivial initial condition {\small $x_{sc}(0),~\dot{x}_{sc}(0) = 0$}, the solutions grow towards infinity as time increases, confirming the physical insight that the DRS-LIP is unstable under the considered operating condition.

\vspace{-0.15 in}
\subsection{Planner Validation}

The efficiency and feasibility of the proposed planner are validated through both simulations and experiments.

\subsubsection{Simulation and experimental setup}

\begin{figure}[t]
    \centering
    \includegraphics[width= 0.9\linewidth]{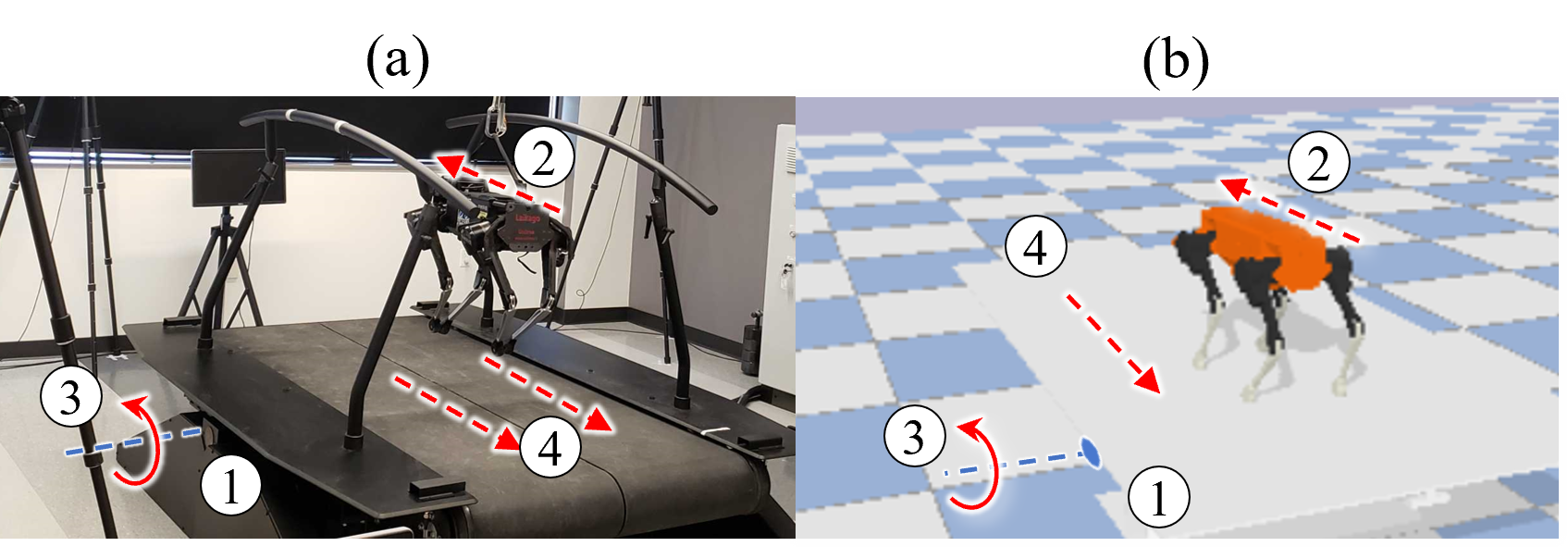}
        \vspace{-0.1 in}
    \caption{Setup of (a) experiments and (b) Pybullet simulations for testing the planner effectiveness under a pitching surface motion.
    The treadmill (\textcircled{1}) has a split belt (\textcircled{4}) that moves at a constant speed while the treadmill rocks about the horizontal axis (\textcircled{3}).
    The robot is a Laikago quadruped (\textcircled{2}).}
    \label{Fig:Simulated_exp_setup}
        \vspace{-0.2 in}
\end{figure}

The validation of the planner utilizes a Laikago quadruped (Fig.~\ref{Fig:Simulated_exp_setup}) developed by Unitree Robotics.
The dimension of the robot is {\small 55} cm $\times$ {\small 35} cm $\times$ {\small 60} cm.
The robot weighs {\small 25} kg in total. 
Each leg weighs {\small 2.9} kg, including its motors that weigh {\small 1.5} kg and are located near the trunk.
It has twelve independently actuated joints, and the torque limits of hip-roll, hip-pitch, and knee-pitch motors are {\small 20} Nm, {\small 55} Nm, and {\small 55} Nm, respectively.

\noindent \textbf{Gait parameters.}
Recall that the proposed planner takes user-defined gait parameters as its input.
To assess the planner under different gait parameters, two sets of parameters (G1) and (G2) are used (see Table II).

\noindent \textbf{DRS motion.}
Three DRS motions are tested to assess the effectiveness of the planner under different surface motions:
\begin{itemize}
    \item [(DRS1)]The DRS motion is vertical and sinusoidal with {\small $A=10$} cm and {\small $\omega=\pi$} rad/s, which emulates vessel motions in regular sea waves~\cite{Vessels_Vertical_Periodic_1954}.
    \item[(DRS2)] The motion consists of a sinusoidal pitching motion with an amplitude of {\small $ 5^{\circ}$} and frequency of {\small $0.5$} Hz.
    \item[(DRS3)] Similar to (DRS2), the motion comprises a sinusoidal pitching motion with an amplitude of {\small $ 5^{\circ}$} and frequency of {\small $0.4$ Hz}.
\end{itemize}
Surface motions (DRS2) and (DRS3) reasonably satisfy the assumption (A3) because the surface's horizontal velocity is approximately constant (specifically, zero) due to the small pitching amplitude.
Still, their accelerations in the vertical direction are relatively significant for validating the proposed method. 
The amplitude of the robot's CoP vertical displacement under (DRS1)-(DRS3) is approximately {\small 10} cm, {\small 7} cm, and {\small 12.5} cm, respectively.

\noindent \textbf{Simulated and physical DRSes.}
To validate the planner feasibility, (DRS1) is implemented in Pybullet simulations alone, and the surface motions (DRS2) and (DRS3) are realized both in simulations and experimentally by a physical Motek M-gait treadmill (see Fig.~\ref{Fig:Simulated_exp_setup}).
During the testing, the robot is placed approximately {\small $100$} cm away from the treadmill's axis of rotation.
The treadmill weighs {\small 750} kg with a dimension of {\small 2.3} m  $\times$ {\small 1.82} m $\times$ {\small 0.5} m.
A {\small 4.5} kW servo motor powers each of of the treadmill's two belts. 
The treadmill can be pre-programmed to perform user-defined pitching (but not vertical) motions and belt translation.
The belt speed is set to be the same as the desired walking speed.
{\small
\begin{table}[]
    \centering
    \caption{User-defined gait parameters in motion planning.}
    \label{table:Planer parameters}
    \vspace{-0.1 in}
    \begin{tabularx}{\columnwidth}{c|X X }
     \hline
      \hline
  \small   Gait parameter     &\small   (G1) &\small  (G2) 
    \\
     \hline
    \hline
     \small Friction coefficient   &\small 0.5 &\small 0.5
     \\
    \small  Robot's base height $z_0$ (cm) 
       &\small 42  &\small 42 
     \\
    \small  Gait duration (s)  &\small  2  &\small  2.5 
     \\
    \small  Average walking velocity (cm/s)   &\small  5  &\small  6 
     \\
     \small Step length (cm)   &\small  10  &\small  15  \\
    \small  Max. step height (cm)    &\small  5  &\small  4 
     \\
     \hline
    \end{tabularx}
    \vspace{-0.25 in}
\end{table}}

\subsubsection{Validation of planner efficiency}
To validate that using the proposed analytical solution improves planner efficiency compared with using the numerical solution, the higher-layer CoM trajectory planning problem is solved based on both solutions under the user-defined gait parameters (G1) and surface motion (DRS2).
For simplicity, the cost function {\small ${h}$} in~\eqref{Optimization_P1} is chosen as trivial. Also, a {\small $6^{th}$}-order B\'ezier curve is used to design the desired swing-foot trajectory for allowing more freedom in trajectory design.

To demonstrate the improved efficiency under different common solvers, both MATLAB and C++ are used to solve the optimization-based planning problem with the same initial guess for {\small 1000} runs.
For fairness of comparison, the optimality and constraint tolerances are set as {\small $ 10^{-6}$} in all runs.
In MATLAB, fmincon is used with an interior-point solver.
For the C++ optimization, the nonlinear optimization solver of the Ipopt package \cite{wachter2006implementation_ipopt} is utilized with the same tolerance setting as MATLAB.


Table \ref{table:Comparision_time} shows that the time cost of the analytical solution based planning is approximately {\small $5$} times shorter than the numerical solution based one.
Since the lower-layer planner is essentially trajectory interpolation, solving it is typically fast (e.g., MATLAB can solve it within {\small $<2$} ms).

Table \ref{table:Comparision_time} also indicates that the higher-layer planner takes {\small $51.4\pm 3.5$} ms to generate the desired CoM trajectory when it is solved by C++ using the approximate anaytical solution.
Therefore, the corresponding time cost for solving both higher and lower layers will be less than {\small $60$} ms, which is much shorter than the gait period of {\small $2$} s.
With such a planning speed,
the planner will be capable of quickly re-planning the desired full-order trajectories in case of any significant changes in the DRS motion.
This planning scheme is reasonable for real-world DRSes, such as vessels~\cite{ShipMotion_tannuri2003estimating}, because the changes of their motions are relatively slow due to low motion frequency ({\small 0.01$\sim$1 Hz}).

\begin{table}[h!]
\vspace{-0.1 in}
\centering
\caption{Average time cost of 1000 runs of higher-layer planning (mean$\pm$SD) under gait parameters (G1) and surface motion (DRS2)}
\vspace{-0.15 in}
\label{table:Comparision_time}
\begin{tabular}{c | c c } 
\hline
\hline
\small Solution method & \small MATLAB  &\small  C++ \\ 
 &\small  (fmincon) &\small  (Ipopt)\\
 \hline\hline 
\small  Numerical (ms)   &\small  $1320.7 \pm 13.8$  &\small  $387.6\pm 19.2$ \\[0ex]  
\small  Analytical (ms)  &\small  $269.1\pm 12.9$  &\small  $51.4\pm 3.5$  \\ [0ex] 
 \hline
\end{tabular}
\vspace{-0.1 in}
\end{table}

\subsubsection{Validation of planner feasibility}
Beside efficiency, the proposed DRS-LIP and its solution also help guarantee planning feasibility.
To test the feasibility of the planned motion,
our previous controller \cite{iqbal2020provably}, which is derived based on input-output linearization and proportional derivative (PD) control, is utilized to track the planned full-order trajectories in Pybullet simulations and experiments.
As this controller does not explicitly ensure the feasibility of ground contact forces, the planned trajectory needs to be physically feasible in order for the controller to be effective.
Thus, if the controller is able to reliably track the planned motion and sustain walking on a DRS, then the physical feasibility of the proposed planner is confirmed.
To help ensure a reasonable tracking performance, PD gains are tuned as {\small $0.7$} and {\small $1.0$} in simulations, and {\small $5.5$} and {\small $0.15$} on hardware.

\begin{figure}[t]
    \centering
    \includegraphics[width= 0.9\linewidth]{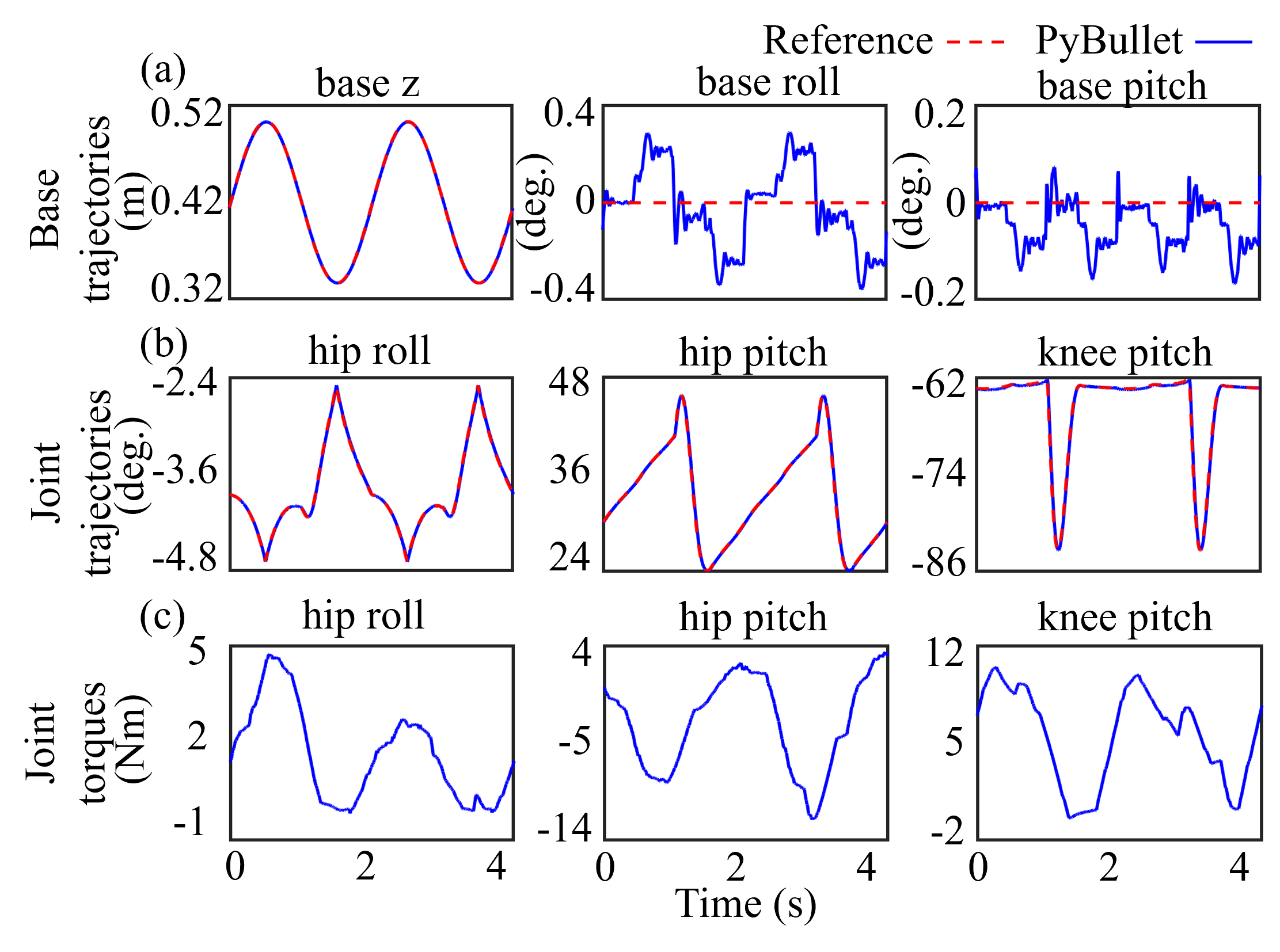}
    \vspace{-0.2 in}
    \caption{Pybullet simulation results at the robot's front-right leg under gait parameters (G1) and surface motion (DRS1).}
    \label{Fig:G1_DRS1_sim}
        \vspace{-0.2 in}
\end{figure}

To validate the planner feasibility under different gait parameters and surface motions, the gait parameters (G1) and the surface motion (DRS1), as well as (G2) and (DRS3), are tested in Pybullet simulations.
Figure \ref{Fig:G1_DRS1_sim} shows the Pybullet simulation results under (G1) and (DRS1).
The robot sustains walking for the entire testing duration, which is over {\small 50} gait cycles.
The base and joint trajectories closely track their reference values, as shown in subplots (a) and (b). 
Also, the actual robot motion respects the torque limits, as indicated in subplot (c).
The simulation results under (G2) and (DRS3) show position tracking performance and torque profiles similar to the case under (G1) and (DRS1), and is given in supplementary material for space consideration.

\begin{figure}[t]
    \centering
    \includegraphics[width= 0.9\linewidth]{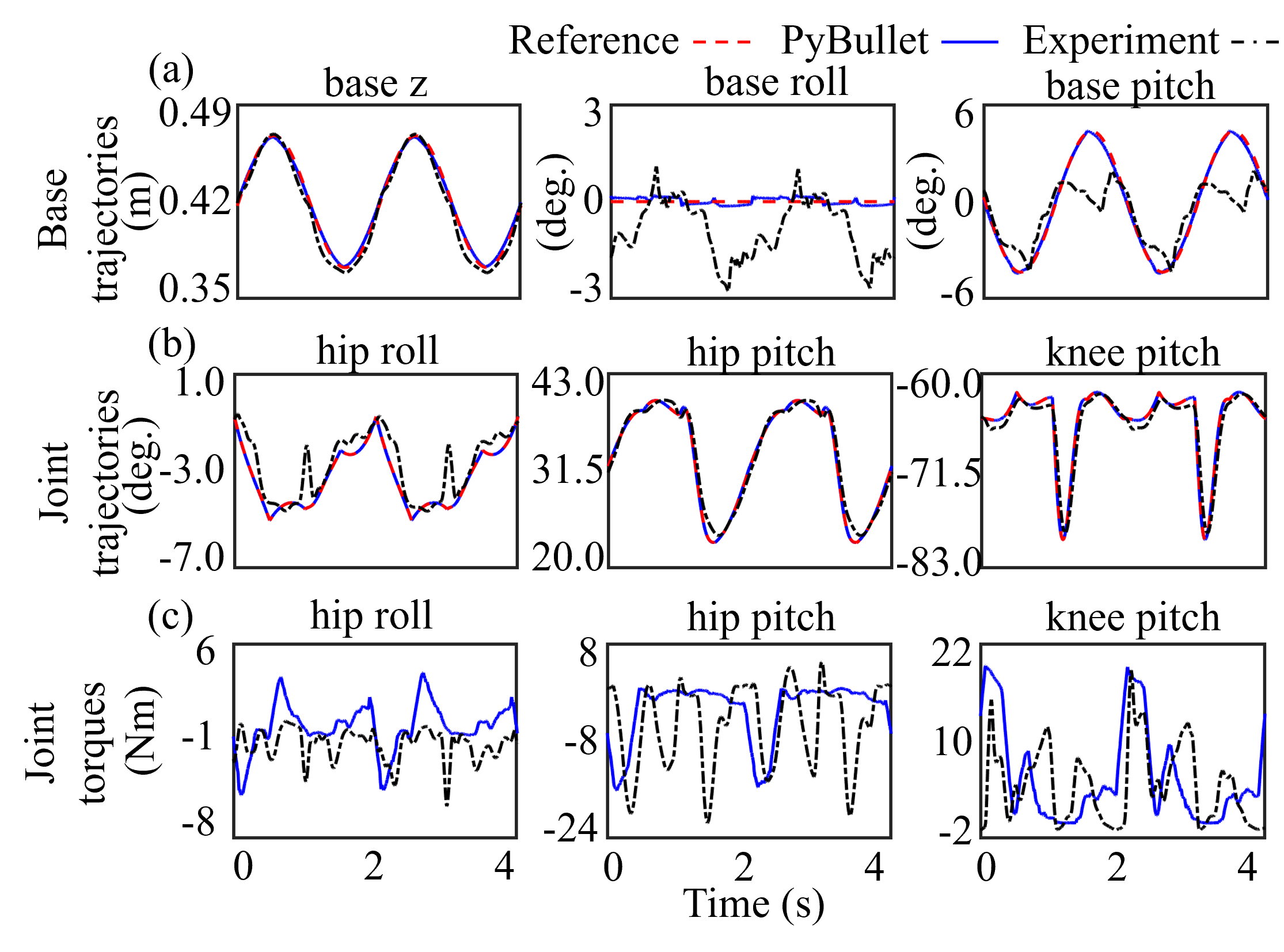}
    \vspace{-0.2 in}
    \caption{Hardware experiment and Pybullet simulation results at the robot's front-right leg under gait parameters (G1) and surface motion (DRS2).}
    \label{Fig:G2_DRS2}
        \vspace{-0.2 in}
\end{figure}

The combination of (G1) and (DRS2) is tested in both simulations and experiments, with the results presented in Fig.~\ref{Fig:G2_DRS2}.
Experiment videos are included in the paper submission, and is also available at \href{https://youtu.be/Amrb3WxLpPs}{\tt \small https://youtu.be/Amrb3WxLpPs}.
In both simulations and experiments, the robot walking is stable, as indicated by the trajectory tracking accuracy in subplots (a) and (b) as well as the experiment video.
Moreover, as displayed in subplot (c), the joint torque limits are met in both the simulation and experiment.
However, the torque profiles of the right leg's three joints show notable discrepancies between Pybullet and experiment results, {which could be caused by the differences between the simulated and actual robot dynamics as well as the different inherent meanings of their effective PD gains}.
Also, from the experiment video, we can see that the robot experiences relatively notable rebounding and slipping at contact switching events when a rear leg lands on the surface.
This violation of the planned contact sequence is essentially due to the robot's temporary loss of contact force feasibility, which could be mitigated through controller design as discussed in Sec.~\ref{sec: discussion}.

%

\vspace{-0.15 in}
\section{DISCUSSION}
\label{sec: discussion}
\vspace{-0.05 in}

This paper has introduced a reduced-order dynamic model of a legged robot that walks on a DRS, by analytically extending the classical LIP model from stationary surfaces~\cite{kajita20013d} to a DRS (e.g., a vessel).
The resulting DRS-LIP model in~\eqref{Eq-LIPM_on_DRS_simplified} is a linear, second-order differential equation, similar to the classical LIP.
However, the DRS-LIP is explicitly time-varying whereas the classical LIP is time-invariant.
This fundamental difference is due to the time-varying movement of the surface at the surface-foot contact points.
This study also provides the stability condition of the DRS-LIP based on the Floquet theory.
Analogous to the classical LIP~\cite{kajita20013d}, the DRS-LIP is unstable under the usual movement range of real-world DRSes such as vessels~\cite{ShipMotion_tannuri2003estimating} (Sec.~\ref{sec: results}-A).

The DRS-LIP is valid under the assumption that the actual robot's rate of whole-body angular momentum about the CoM is negligible (assumption (A1)).
To relax this assumption, the point mass of the proposed DRS-LIP could be augmented with a fly wheel~\cite{pratt2006capture,zhao2017robust} to account for nonzero rate of angular momentum.
Also, the DRS-LIP can be generalized from a constant CoM height (as enforced by assumption (A2)) to a varying height by integrating with the variable-height LIP for stationary surfaces~\cite{LIP_varyingHeight_caron2020biped}.

This study also derives the approximate analytical solution of the DRS-LIP 
for vertical, sinusoidal surface motions.
Its improved accuracy and computational efficiency compared with numerical solutions are confirmed through MATLAB simulations, as shown in Fig.~\ref{Fig-SolAccuracy} and Table~\ref{table:Comparision_time_Sol}.
Yet, the solution is not valid for periodic surface motions that are not strictly sinusoidal or vertical.
Under general periodic surface motions, the DRS-LIP in \eqref{Eq-LIPM_on_DRS_simplified} becomes Hill's equation~\cite{farkas2013periodic}, and thus the solution could be derived based on existing results of Hill's equations.
Also, as indicated by \eqref{Eq:simplified_Cap-xy}, the DRS-LIP could become nonhomogeneous under nontrivial horizontal DRS motions, with forcing terms induced by the horizontal DRS accelerations.
The effects of horizontal DRS motions on robot's movement could be studied based on \eqref{Eq:simplified_Cap-xy}.

To highlight the usefulness of the analytical results, they have been used as a basis to synthesize a hierarchical planner that efficiently produces desire, physically feasible motions for quadrupedal robot walking on a DRS.
The feasibility of the planned motion is validated by using our previous trajectory tracking controller~\cite{iqbal2020provably} to command a quadrupedal robot to follow the planned motion during DRS walking.
As discussed in Sec.V-B, 
both simulation and experimental results indicate the reasonable feasibility of the proposed planner under different gait parameters and surface motions (see Figs \ref{Fig:G1_DRS1_sim} and \ref{Fig:G2_DRS2}).
To mitigate the temporary violation of the planned gait sequence observed in experiments, which is partly induced by the discrepancies between the DRS-LIP and the actual robot dynamics, the planned motion could be tracked by an optimization-based controller that explicitly ensures physical feasibility~\cite{iqbal2021extended}.

\vspace{-0.15 in}
\section{CONCLUSION}
\vspace{-0.05 in}
\label{sec: conclusion}
This paper has introduced a reduced-order model that captures the essential dynamic behavior of a legged robot that walks on a nonstationary rigid surface (i.e., a DRS).
The model was derived based on the theoretical extension of the classical LIP model from a stationary surface to a DRS.
The model was found to be unstable under the common operating conditions of a real-world DRS (e.g., a vessel in regular sea waves).
Its approximate analytical solution was also provided and used as a basis to develop a motion planner that efficienctly generates feasible robot motions for quadrupedal walking on a DRS.
The stability property of the model and the efficiency and accuracy of its solution were validated through MATLAB and C++ simulations.
The effectiveness of the proposed planner was demonstrated through both physics-based simulations and experiments on a Laikago quadrupedal robot under different gait parameters and surface motion profiles.

\vspace{-0.1 in}
\section*{Acknowledgments}
\vspace{-0.05 in}
The authors would like to thank Tingnan Zhang and Jie Tan for supporting the experimental validation.

    \vspace{-0.1 in}
\bibliography{ReferencesAbbrev} 

\begin{thebibliography}{10}

\bibitem{anymal_16}
M.~Hutter, C.~Gehring, D.~Jud, A.~Lauber, C.~D. Bellicoso, V.~Tsounis,
  J.~Hwangbo, K.~Bodie, P.~Fankhauser, M.~Bloesch, {\em et~al.}, ``{ANYmal-a}
  highly mobile and dynamic quadrupedal robot,'' in {\em Proc. IEEE/RSJ Int.
  Conf. Intel. Robot. Syst. (IROS)}, pp.~38--44, 2016.

\bibitem{fawcett2021robust}
R.~T. Fawcett, A.~Pandala, A.~D. Ames, and K.~A. Hamed, ``Robust stabilization
  of periodic gaits for quadrupedal locomotion via qp-based virtual constraint
  controllers,'' {\em IEEE Contr. Syst. L.}, vol.~6, pp.~1736--1741, 2021.

\bibitem{zhang2021efficient}
Z.~Zhang, J.~Yan, X.~Kong, G.~Zai, and Y.~T. Liu, ``Efficient motion planning
  based on kinodynamic model for quadruped robots following persons in confined
  spaces,'' {\em IEEE/ASME Trans. Mechatron.}, 2021.

\bibitem{winkler2015planningSemini}
A.~W. Winkler, C.~Mastalli, I.~Havoutis, M.~Focchi, D.~G. Caldwell, and
  C.~Semini, ``Planning and execution of dynamic whole-body locomotion for a
  hydraulic quadruped on challenging terrain,'' in {\em Proc. IEEE Int. Conf.
  Rob. Autom.}, pp.~5148--5154, 2015.

\bibitem{iqbal2020provably}
A.~Iqbal, Y.~Gao, and Y.~Gu, ``Provably stabilizing controllers for quadrupedal
  robot locomotion on dynamic rigid platforms,'' {\em IEEE/ASME Trans.
  Mechatron.}, vol.~25, no.~4, pp.~2035--2044, 2020.

\bibitem{gao2021invariant}
Y.~Gao, C.~Yuan, and Y.~Gu, ``Invariant extended kalman filtering for hybrid
  models of bipedal robot walking,'' in {\em Proc. IFAC Mod. Est. Contr.
  Conf.}, vol.~54, pp.~290--297, 2021.

\bibitem{motahar2016composing}
M.~S. Motahar, S.~Veer, and I.~Poulakakis, ``Composing limit cycles for motion
  planning of 3d bipedal walkers,'' in {\em Proc. IEEE Conf. Dec. Contr.},
  pp.~6368--6374, 2016.

\bibitem{kajita20013d}
S.~Kajita, F.~Kanehiro, K.~Kaneko, K.~Yokoi, and H.~Hirukawa, ``The 3d linear
  inverted pendulum mode: A simple modeling for a biped walking pattern
  generation,'' in {\em Proc. IEEE Int. Conf. Intel. Robot. Sys.}, vol.~1,
  pp.~239--246, 2001.

\bibitem{pratt2006capture}
J.~Pratt, J.~Carff, S.~Drakunov, and A.~Goswami, ``Capture point: A step toward
  humanoid push recovery,'' in {\em Proc. IEEE-RAS Int. Conf. Humanoid Robot.},
  pp.~200--207, 2006.

\bibitem{zhao2017robust}
Y.~Zhao, B.~R. Fernandez, and L.~Sentis, ``Robust optimal planning and control
  of non-periodic bipedal locomotion with a centroidal momentum model,'' {\em
  Int. J. Rob. R.}, vol.~36, no.~11, pp.~1211--1242, 2017.

\bibitem{mastalli2020motion}
C.~Mastalli, I.~Havoutis, M.~Focchi, D.~G. Caldwell, and C.~Semini, ``Motion
  planning for quadrupedal locomotion: Coupled planning, terrain mapping, and
  whole-body control,'' {\em IEEE Trans. Robot.}, vol.~36, no.~6,
  pp.~1635--1648, 2020.

\bibitem{Kajita2003BipedWP}
S.~Kajita, F.~Kanehiro, K.~Kaneko, K.~Fujiwara, K.~Harada, K.~Yokoi, and
  H.~Hirukawa, ``Biped walking pattern generation by using preview control of
  zero-moment point,'' in {\em Proc. IEEE Int. Conf. Robot. Autom.}, vol.~2,
  pp.~1620--1626, 2003.

\bibitem{westervelt2007feedback}
E.~R. Westervelt, J.~W. Grizzle, C.~Chevallereau, J.~H. Choi, and B.~Morris,
  {\em Feedback control of dynamic bipedal robot locomotion}, vol.~28.
\newblock CRC press, 2007.

\bibitem{gong2020angular}
Y.~Gong and J.~Grizzle, ``Angular momentum about the contact point for control
  of bipedal locomotion: Validation in a lip-based controller,'' {\em arXiv
  preprint arXiv:2008.10763}, 2020.

\bibitem{LIP_varyingHeight_caron2020biped}
S.~Caron, ``Biped stabilization by linear feedback of the variable-height
  inverted pendulum model,'' in {\em Proc. IEEE Int. Conf. Robot. Autom.},
  pp.~9782--9788, 2020.

\bibitem{xiong2020global}
X.~Xiong, J.~Reher, and A.~Ames, ``Global position control on underactuated
  bipedal robots: Step-to-step dynamics approximation for step planning,'' {\em
  arXiv preprint arXiv:2011.06050}, 2020.

\bibitem{nagarajan2014balancing}
U.~Nagarajan and K.~Yamane, ``Balancing in dynamic, unstable environments
  without direct feedback of environment information,'' {\em IEEE Trans.
  Robot.}, vol.~30, no.~5, pp.~1234--1241, 2014.

\bibitem{Yamenzheng2011ball}
Y.~Zheng and K.~Yamane, ``Ball walker: A case study of humanoid robot
  locomotion in non-stationary environments,'' in {\em Proc. IEEE Int. Conf.
  Robot. Autom.}, pp.~2021--2028, 2011.

\bibitem{BallMan2020Koshil_yang}
C.~{Yang}, B.~{Zhang}, J.~{Zeng}, A.~{Agrawal}, and K.~{Sreenath}, ``Dynamic
  legged manipulation of a ball through multi-contact optimization,'' in {\em
  Proc. IEEE/RSJ Int. Conf. Intel. Robot. Syst.}, pp.~7513--7520, 2020.

\bibitem{asano2021modeling}
F.~Asano, ``Modeling and control of stable limit cycle walking on floating
  island,'' in {\em Proc. IEEE Int. Conf. Mechatron.}, pp.~1--6, 2021.

\bibitem{iqbal_SLIP}
A.~Iqbal, Z.~Mao, and Y.~Gu, ``Modeling, analysis, and control of slip running
  on dynamic platforms,'' {\em ASME L. Dyn. Sys. Contr.}, vol.~1, no.~2, 2021.

\bibitem{kapitza1951dynamic}
P.~Kapitza, ``Dynamic stability of a pendulum with an oscillating point of
  support,'' {\em Zh. Eksp. Teor. Fiz}, vol.~21, p.~588, 1951.

\bibitem{iqbal2021extended}
A.~Iqbal and Y.~Gu, ``Extended capture point and optimization-based control for
  quadrupedal robot walking on dynamic rigid surfaces,'' in {\em Proc. IFAC
  Mod. Est. Contr. Conf.}, vol.~54, pp.~72--77, 2021.

\bibitem{Ship_Motion_Monitoring_System}
H.-K. Yoon, G.-J. Lee, and D.-K. Lee, ``Development of the motion monitoring
  system of a ship,'' {\em J. Navig. Port Res.}, vol.~32, no.~1, pp.~15--22,
  2008.

\bibitem{gahlinger2000const_hv}
P.~M. Gahlinger, ``Cabin location and the likelihood of motion sickness in
  cruise ship passengers,'' {\em J. Trav. Med.}, vol.~7, no.~3, pp.~120--124,
  2000.

\bibitem{Vessels_Vertical_Periodic_1954}
T.~B. Benjamin and F.~J. Ursell, ``The stability of the plane free surface of a
  liquid in vertical periodic motion,'' {\em Proc. Roy. Soc. London.},
  vol.~225, no.~1163, pp.~505--515, 1954.

\bibitem{farkas2013periodic}
M.~Farkas, {\em Periodic motions}.
\newblock Springer, 2013.

\bibitem{phelps1965analytical}
F.~Phelps~III and J.~Hunter~Jr, ``An analytical solution of the inverted
  pendulum,'' {\em Amer. J. Phys.}, vol.~33, no.~4, pp.~285--295, 1965.

\bibitem{BookB_ME_werth2005charged}
F.~Werth, N.~Gheorghe, F.~Major, V.~Gheorghe, G.~Werth, S.~Major, and G.~Werth,
  {\em Charged Particle Traps: Physics and Techniques of Charged Particle Field
  Confinement}.
\newblock Springer Ser. Atom., Opt., Plas. Phys., Springer, 2005.

\bibitem{Hills_det_Simplification_bateman1953higher}
H.~Bateman, {\em Higher transcendental functions [volumes i-iii]}, vol.~1.
\newblock McGraw-Hill Book Company, 1953.

\bibitem{ShipMotion_tannuri2003estimating}
E.~A. Tannuri, J.~V. Sparano, A.~N. Simos, and J.~J. Da~Cruz, ``Estimating
  directional wave spectrum based on stationary ship motion measurements,''
  {\em App. Ocean Res.}, vol.~25, no.~5, pp.~243--261, 2003.

\bibitem{wachter2006implementation_ipopt}
A.~W{\"a}chter and L.~T. Biegler, ``On the implementation of an interior-point
  filter line-search algorithm for large-scale nonlinear programming,'' {\em
  Math. Prog.}, vol.~106, no.~1, pp.~25--57, 2006.

\end{thebibliography}
\bibliographystyle{ieeetr}





\ifCLASSOPTIONcaptionsoff
  \newpage
\fi

\end{document}


\maketitle    


\section{Introduction}

This document contains the supplementary material for the paper entitled ``DRS-LIP: Linear Inverted Pendulum Model for Legged Locomotion on Dynamic Rigid Surfaces". 

\section{Derivation Details of Proposed Analytical Approximate Solution}

This section gives the derivation details for the proposed analytical approximate solution that are omitted in Sec. III of the main paper.


\subsection{Recurrence Relationship between $\mu$ and $\beta_n$}
This subsection presents the derivation of the recurrence relationship between $\mu$ and $\beta_n$, which is given in Eq. (9) in the main paper.
The original derivation is given in the reference [29] cited in the main paper.

Mathieu's equation formed in this study (i.e., Eq. (7) in the main paper) has the following form:
\begin{equation}
    \frac{d^2x_{sc}}{d\tau^2} +(c_0 - 2c_1\cos 2 \tau)x_{sc}  = 0.
    \label{Eq-ME_A}
\end{equation}

According to the reference [29] cited in the main paper, an assumed solution satisfying \eqref{Eq-ME_A} is given by:
\begin{equation}
    {x}_{sc}(\tau) = e^{\mu \tau}\sum_{n=-\infty}^{\infty}C_{2n} e^{i2n\tau}.
    \label{Eq-Assumed_Sol1}
\end{equation}
For such a solution to exist, it must satisfy \eqref{Eq-ME_A}.
Taking the double derivative of the assumed solution in \eqref{Eq-Assumed_Sol1} and substituting it back in \eqref{Eq-ME_A}, we have:
\begin{equation}
    \sum_{n=-\infty}^{\infty}[(\mu^2 +(i2n)^2 +4i\mu n)C_{2n} + (c_0 -2c_1 \cos 2 \tau)C_{2n}] e^{(i2n+\mu)\tau} =0.
    \label{Eq-RR1}
\end{equation}

By using Euler's formula we can write $\cos 2 \tau =\frac{e^{i2\tau}+e^{-i2\tau}}{2}$, which can be substituted in
\eqref{Eq-RR1} to yield:
\begin{equation}
\begin{aligned}
    \sum_{n=-\infty}^{\infty}[(\mu^2 &-(2n)^2 +2 (i\mu) (2n)+c_0)C_{2n} + \\ &-c_1(e^{i2\tau}+e^{-i2\tau})C_{2n}] e^{(i2n+\mu)\tau} =0.
    \label{Eq-RR2}\\
\end{aligned}
\end{equation}

Rewriting \eqref{Eq-RR2} gives:
\begin{equation}
\begin{aligned}
    \sum_{n=-\infty}^{\infty}[((i\mu)^2 &+(2n)^2 -2 (i\mu) (2n)-c_0)C_{2n}e^{(i2n+\mu)\tau} + \\ &c_1(C_{2n}e^{(i2(n+1)+\mu)\tau}+C_{2n}e^{(i2(n-1)+\mu)\tau}]  =0.
    \label{Eq-RR3}\\
\end{aligned}
\end{equation}

Since \eqref{Eq-RR3} is valid for infinite terms, we define new indices $l:=n+1$ and $m:=n-1$ and replace $n$ with them to obtain the following equations:
\begin{equation}
\begin{aligned}
    &\sum_{n=-\infty}^{\infty}[((2n-i\mu)^2-c_0)C_{2n}+ c_1C_{2(n+1)}+c_1C_{2(n-1)}]e^{(i2n+\mu)\tau}   =0,\\
    &\Rightarrow \frac{c_1}{((2n-i\mu)^2-c_0)}C_{2(n+1)} + C_{2n} + \frac{c_1}{((2n-i\mu)^2-c_0)}C_{2(n-1)} =0,\\
    &\Rightarrow \beta_n(\mu) C_{2(n+1)} + C_{2n} + \beta_n(\mu) C_{2(n-1)} =0,    
    \label{Eq-RR4}\\
\end{aligned}
\end{equation}
where $\beta_n(\mu) : = \frac{c_1}{((2n-i\mu)^2-c_0)}$. 


\subsection{Computation of solution coefficient $C_{2n}$}
The detailed derivation of solution coefficient $C_{2n}$, which is omitted in Sec.III-A-3 of the main paper, is presented here and is originally given in reference [29] cited in the main paper.

Upon analyzing the recurrence relation in \eqref{Eq-RR4}, we see that for sufficiently large $n$ (e.g., $n>N$) the coefficient satisfies $C_{2n}<<C_{2(n-1)}$.
Accordingly, coefficients at higher indices can be neglected (i.e., $C_{2(n+1)}=0$).
Thus, solving the recurrence relation for various indices we get:
\begin{equation}
\begin{aligned}
    &\text{for}~n = N:\\
    & \beta_N C_{2(N+1)} + C_{2N} + \beta_N C_{2(N-1)} =0, \\ 
    &\Rightarrow C_{2N} = - \beta_N C_{2(N-1)},~\text{since} ~C_{2(N+1)}=0
    \label{Eq-RR5}
\end{aligned}
\end{equation}

\begin{equation}
\begin{aligned}
    &\text{for}~n = N-1:\\
    & \beta_{N-1} C_{2N} + C_{2(N-1)} + \beta_{N-1} C_{2(N-2)} =0, \\ 
    &\Rightarrow C_{2(N-1)} = \frac{- \beta_{N-1} C_{2(N-2)}}{1- \beta_N \beta_{N-1}}
    \label{Eq-RR6}
\end{aligned}
\end{equation}

\begin{equation}
\begin{aligned}
    &\text{for}~n = N-2:\\
    & \beta_{N-2} C_{2(N-1)} + C_{2(N-2)} + \beta_{N-2} C_{2(N-3)} =0, \\ 
    &\Rightarrow C_{2(N-2)} = \frac{- \beta_{N-2} C_{2(N-3)}}{1- \frac{ \beta_{N-2} \beta_{N-1}}{1- \beta_{N-1}\beta_N }}
    \label{Eq-RR7}
\end{aligned}
\end{equation}

Accordingly, we can write a general relation for computing the coefficient $C_{2n}$ as:
\begin{equation}
\begin{aligned}
     C_{2n} = \frac{- \beta_{n} C_{2(n-1)}}{1- \frac{ \beta_{n}\beta_{n+1} }{1- \frac{\beta_{n+1} \beta_{n+2}}{1-\frac{\beta_{n+2} \beta_{(n+3)}}{1- \frac{\beta_{n+3} \beta_{n+4}}{1-~~.....}}}}}
    \label{Eq-Coefficient}
\end{aligned}
\end{equation}
By setting {$C_0 =A$}
in \eqref{Eq-Coefficient}, all other coefficients can be determined using \eqref{Eq-Coefficient}.
Furthermore, the relation in \eqref{Eq-Coefficient} can be inverted (by replacing index with its additive inverse) to find the coefficient at negative indices.
Recall that $\beta_n$ is defined as $\beta_n(\mu) = \frac{c_1}{((2n-i\mu)^2-c_0)}$.
By this definition, we can see that $\beta_{-n}$ is the complex conjugate of $\beta_n$. 
Therefore, $C_{-2n}$ is the complex conjugate of $C_{2n}$.
That is, if we denote $C_{2n}= r_{2n}e^{i\theta_{2n}}$ with real numbers $r_{2n}$ and $\theta_{2n}$, then $ C_{-2n} =r_{2n}e^{-i\theta_{2n}}$. 
In the next subsection, we utilize these expressions to simplify the expression of the approximate analytical solution for computing the coefficients $\alpha_1$ and $\alpha_2$ under a given initial condition.

\subsection{ Computing $\alpha_1$ and $\alpha_2$ for a given initial condition}
Substituting the complex notation of $C_{2n}$ (i.e., $C_{2n}= r_{2n}e^{i\theta_{2n}}$)
in the approximate analytical solution of Mathieu's equation (i.e., Eq. (11) in the main paper), we have:
\begin{equation}
\begin{aligned}
\small
    \hat{x}_{sc}(\tau) &= \alpha_1 e^{\mu \tau}\sum_{n=-N}^{N} r_{2n}e^{i\theta_{2n}} e^{i2n\tau} +\alpha_2 e^{-\mu \tau}\sum_{n=-N}^{N} r_{2n}e^{i\theta_{2n}} e^{-i2n\tau}\\
    &=\alpha_1 e^{\mu \tau}\sum_{n=-N}^{N} r_{2n} e^{i(2n\tau +\theta_{2n})} +\alpha_2 e^{-\mu \tau}\sum_{n=-N}^{N} r_{2n} e^{-i(2n\tau-\theta_{2n})}\\
    &=\alpha_1 e^{\mu \tau}\sum_{n=1}^{N}[r_{0}+ r_{2n}( e^{i(2n\tau +\theta_{2n})}+e^{-i(2n\tau +\theta_{2n})})] +
    \alpha_2 e^{-\mu \tau}\sum_{n=1}^{N}[r_{0}+ r_{2n}( e^{-i(2n\tau-\theta_{2n})}+e^{i(2n\tau-\theta_{2n}}))]\\
    &=\alpha_1 e^{\mu \tau}\sum_{n=1}^{N}[r_{0}+ 2r_{2n}\cos(2n\tau +\theta_{2n})] +
     \alpha_2 e^{-\mu\tau}\sum_{n=1}^{N}[r_{0}+ 2r_{2n}\cos(2n\tau-\theta_{2n})].
    \label{Eq-Sol_r_th_1}
\end{aligned}
\end{equation}

Transforming the expression in \eqref{Eq-Sol_r_th_1} by replacing $\tau$ with $\frac{\frac{\pi}{2}+\omega t}{2}$ (recall $\tau :=\frac{\frac{\pi}{2}+\omega t}{2}$), we have:
\begin{equation}
\begin{aligned}
\small
    \hat{x}_{sc}(t)
    &=\alpha_1 e^{\mu \frac{\frac{\pi}{2}+\omega t}{2}}\sum_{n=1}^{N}[r_{0}+ 2r_{2n}\cos(\frac{n \pi}{2}+n\omega t +\theta_{2n})] + \alpha_2 e^{-\mu\frac{\frac{\pi}{2}+\omega t}{2}}\sum_{n=1}^{N}[r_{0}+ 2r_{2n}\cos(\frac{n\pi}{2}+n \omega t-\theta_{2n})].
    \label{Eq-Sol_r_th_2}
\end{aligned}
\end{equation}
Given initial conditions $\hat{x}_{sc}(0)$ and $\dot{\hat{x}}_{sc}(0)$, we can solve the coefficients $\alpha_1$ and $\alpha_2$ based on the solution in \eqref{Eq-Sol_r_th_2} and its first derivative computed using \eqref{Eq-Sol_r_th_2}.

\section{Simulation Validation under Gait Parameters (G2) and Surface Motion (DRS2)}
\begin{figure}[t]
    \centering
    \includegraphics[width= 0.6\linewidth]{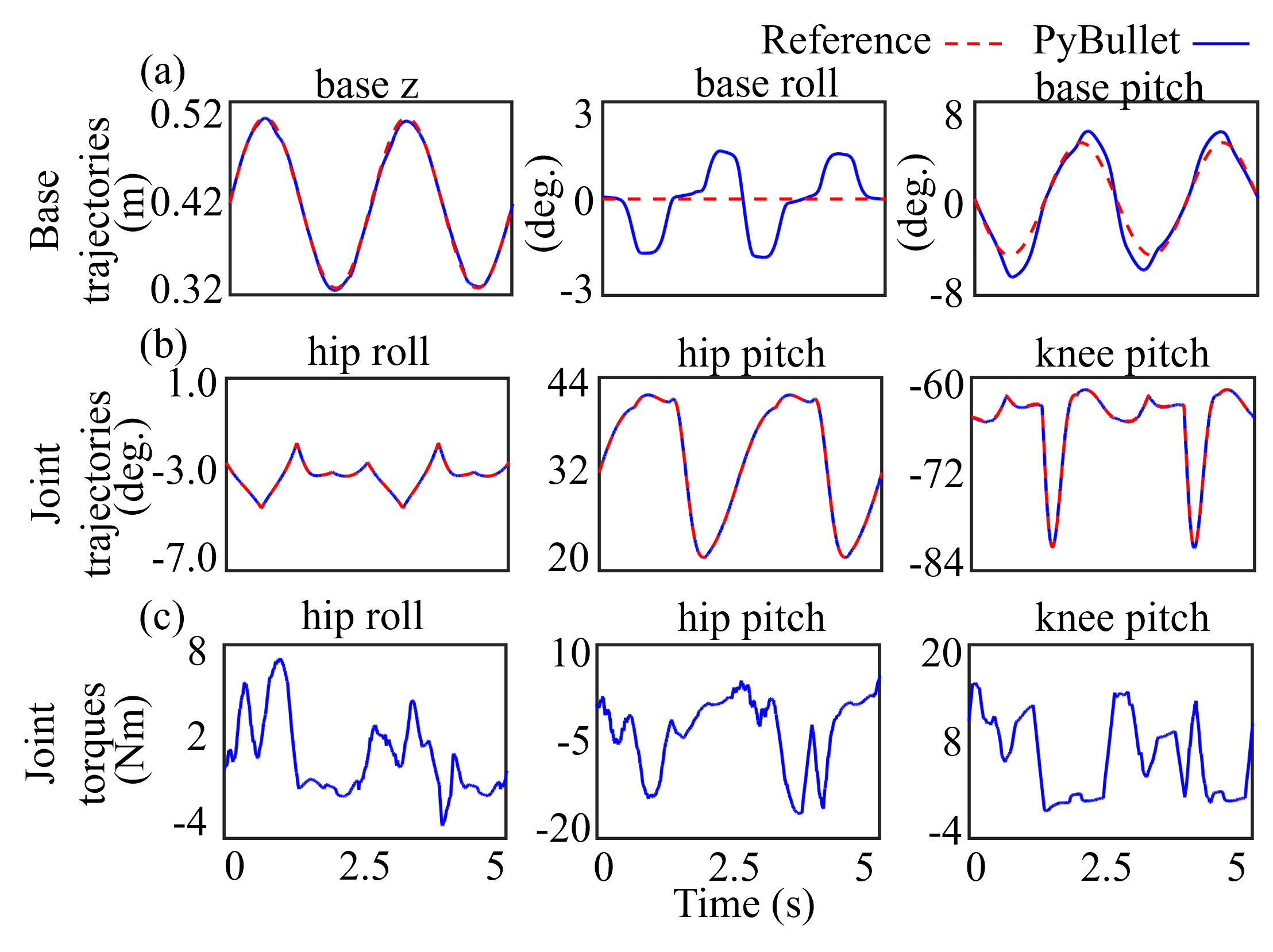}
    \caption{Pybullet simulation results at the robot's front-right leg under gait parameters (G2) and surface motion (DRS3).}
    \label{Fig:G2_DRS3_sim}
\end{figure}

As presented in Sec.V-B-3 of the main paper, three combinations of gait parameters and surface motions are tested to validate the feasibility of the desired full-order position trajectories produced by the proposed planner.
These combinations are: parameter (G1) and motion (DRS1), (G1) and (DRS2), and (G3) and (DRS2).
The results from the first two cases are shown in Figs. 7 and 8 in the main paper.
The Pybullet simulation results from the last case is shown in Fig. \ref{Fig:G2_DRS3_sim} of this supplementary material.
Similar to the first case, the position and orientation tracking of base and swing foot trajectories is accurate (subplots (a) and (b)), and the torque limit is respected (subplot (c)).
Thus, the robot is able to satisfactorily track the planned motion under a controller that does not explicitly ensure gait feasibility, indicating the feasibility of the planned motion.
Recordings of the Pybullet simulation are included in the video submission.


